\RequirePackage[OT1]{fontenc}

\documentclass[journal]{IEEEtran}
\usepackage{ifthen}
\newcommand\arxivmode{true} 
\ifthenelse{\equal{\arxivmode}{true}}{\pdfoutput=1}{}
\usepackage{times}
\usepackage{textcomp}
\usepackage{ragged2e}

\usepackage[numbers]{natbib}
\usepackage{multicol}
\usepackage[backref=page]{hyperref}
\usepackage{amsmath}
\usepackage{amssymb}
\usepackage{graphicx}
\usepackage[numbers]{natbib}
\usepackage{array}
\usepackage{algorithm}
\usepackage[noend]{algpseudocode}
\usepackage{float}
\usepackage[capitalise]{cleveref}
\usepackage[binary-units=true]{siunitx}
\usepackage{subfig}
\usepackage{balance}
\usepackage{fancyhdr}

\usepackage{xcolor}

\usepackage{pgfplots}

\usepackage{pgfplotstable}

\usepackage{tikz}
\usetikzlibrary{external,positioning}
\usepgfplotslibrary{external}
\usepackage{pdfpages}

\newcommand*{\preamblepath}{./support}
\input{\preamblepath/macro.sty}

\RequirePackage{luatex85}

\usepackage{adjustbox}

\usepackage{mwe}

\usepackage{makecell,multirow,booktabs}

\usepackage[utf8]{inputenc}

\newcommand\vizfbox{true}
\newcommand\vizbox[2]{\ifthenelse{\equal{#1}{false}}{\fbox{#2}}{#2}}

\renewcommand*{\backref}[1]{}
\renewcommand*{\backrefalt}[4]{
    \ifcase #1
          \or [Page~#2]
          \else [Pages~#2]
    \fi
}

\fancypagestyle{firstpagestyle}
{
  \fancyhf{}
  \fancyhead[L]{\footnotesize{%
      \begin{center}
          This paper has been accepted for publication in \emph{IEEE Transactions on Robotics}. \\ DOI: \href{https://doi.org/10.1109/TRO.2021.3104459}{10.1109/TRO.2021.3104459} \\
        \end{center}
    }}
  \fancyfoot[L]{\footnotesize{%
      \vspace{-0.5cm}
        \textcopyright 2021 IEEE. Personal use of this material is
        permitted. Permission from IEEE must be obtained for all other uses,
        in any current or future media, including reprinting/republishing this
        material for advertising or promotional purposes, creating new
        collective works, for resale or redistribution to servers or lists, or
        reuse of any copyrighted component of this work in other works.
  }}

}

\begin{document}

\title{Autonomous Cave Surveying with an Aerial Robot}
\author{Wennie~Tabib,
        Kshitij~Goel,
        John~Yao,
        Curtis~Boirum,
        and~Nathan~Michael
        \thanks{The authors are with the Robotics Institute, Carnegie Mellon University,
          Pittsburgh, PA, 15213 USA e-mail: \texttt{\{wtabib, kgoel1, johnyao,
            cboirum, nmichael\}@andrew.cmu.edu}.}
        \ifthenelse{\equal{\arxivmode}{true}}%
        {}%
        {\thanks{Manuscript received March 24, 2020.}}}

\ifthenelse{\equal{\arxivmode}{true}}%
{}%
{\markboth{IEEE Transactions on Robotics}
{Shell \MakeLowercase{\textit{et al.}}: Autonomous Cave Surveying with an Aerial Robot}}%

\maketitle

\begin{abstract}
  This paper presents a method for cave surveying
  in total darkness using an autonomous aerial vehicle equipped with a depth camera for
  mapping, downward-facing camera for state estimation, and forward and
  downward lights. Traditional methods of cave surveying are labor-intensive
  and dangerous due to the risk of hypothermia when collecting data over
  extended periods of time in cold and damp environments, the risk of injury
  when operating in darkness in rocky or muddy environments, and
  the potential structural instability of the subterranean environment.
  Although these dangers can be mitigated by deploying robots to map dangerous passages
    and voids, real-time feedback is often needed to operate robots safely and efficiently.
  Few state-of-the-art, high-resolution perceptual modeling techniques
attempt to reduce their high bandwidth requirements to work
well with low bandwidth communication channels.
  To bridge this gap in the state of
  the art, this work compactly represents sensor
  observations as Gaussian mixture models and maintains a local occupancy grid
  map for a motion planner that greedily maximizes an
  information-theoretic objective function. The approach accommodates both
  limited field of view depth cameras and larger field of view LiDAR sensors and is extensively
evaluated in long duration simulations on an embedded PC. An aerial system
  is leveraged to demonstrate the repeatability of the approach in a flight arena
  as well as the effects of communication dropouts. Finally, the
  system is deployed in Laurel Caverns, a commercially owned and operated cave in
  southwestern Pennsylvania, USA, and a wild cave in West Virginia, USA.
  Videos of the simulation and hardware results are available at
  \href{https://youtu.be/iwi3p7IENjE}{https://youtu.be/iwi3p7IENjE}
  and \url{https://youtu.be/H8MdtJ5VhyU}.
\end{abstract}


%
\IEEEpeerreviewmaketitle
\ifthenelse{\equal{\arxivmode}{true}}%
{\thispagestyle{firstpagestyle}}%
{\markboth{IEEE Transactions on Robotics}
{Shell \MakeLowercase{\textit{et al.}}: Autonomous Cave Surveying with an Aerial Robot}}%

\section{Introduction\label{sec:intro}}
\begin{figure}[t]
  \centering
      \subfloat[\label{sfig:robot_mapping}]{%
        \ifthenelse{\equal{\arxivmode}{true}}%
        {\includegraphics[width=\linewidth,trim=0 0 0 50,clip]{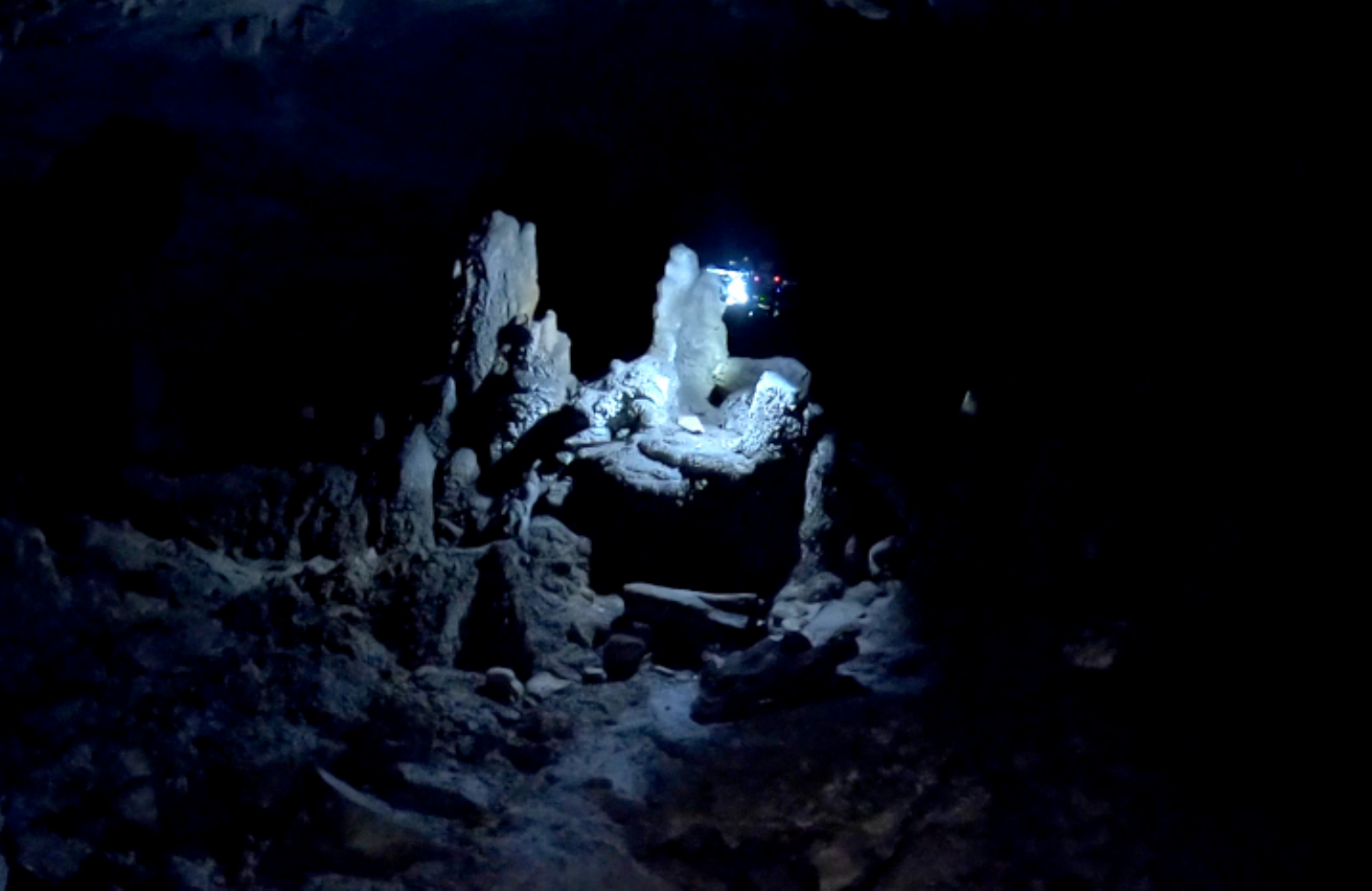}}%
        {\vizbox{\vizfbox}\includegraphics[width=\linewidth,trim=0 0 0 50,clip]{rapps_cave/images/rapps0.eps}}%
      }\\
      \subfloat[\label{sfig:rapps_color_camera}]{%
        \ifthenelse{\equal{\arxivmode}{true}}%
        {\includegraphics[height=1.85cm,trim=70 50 70 70,clip]{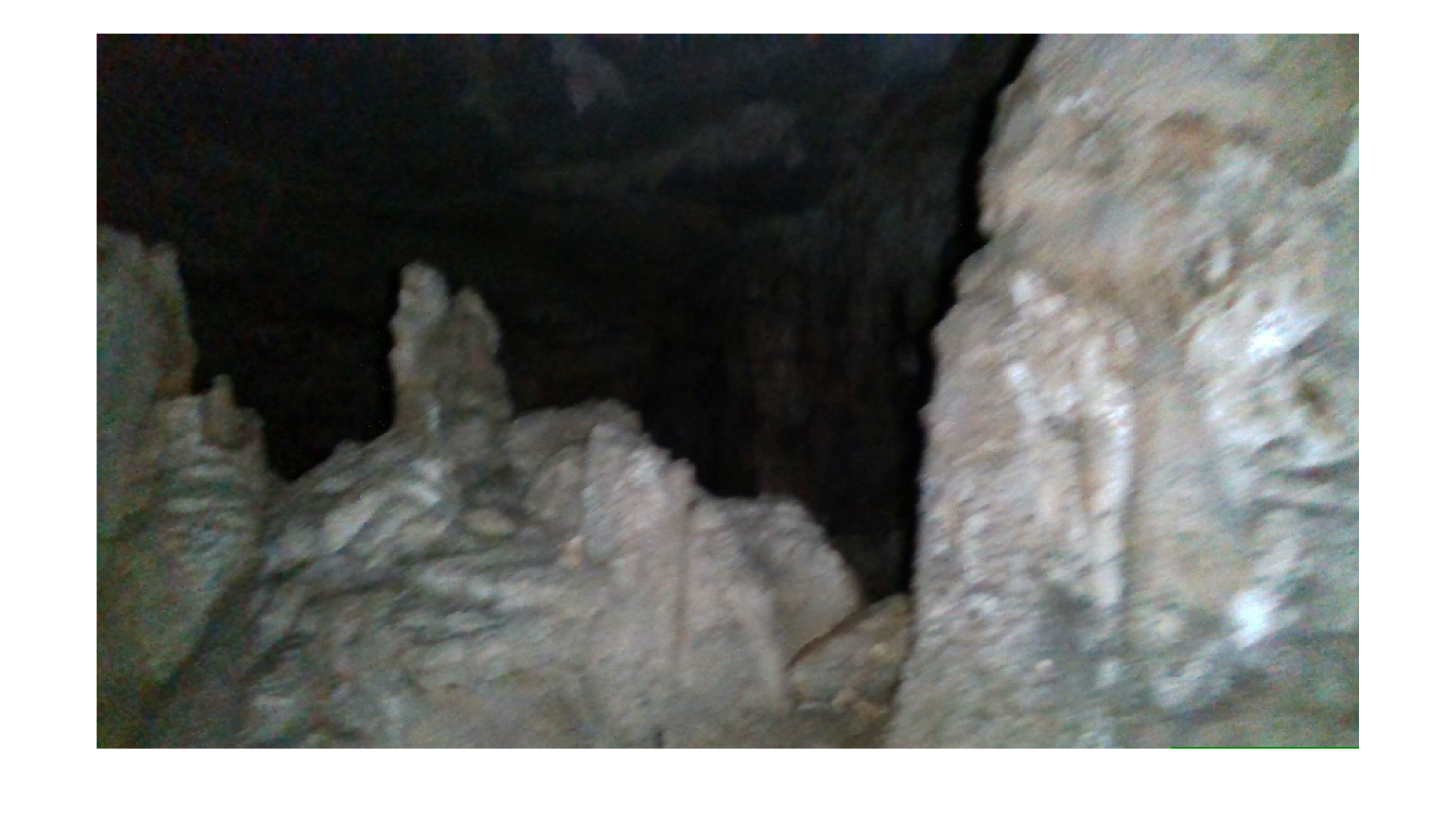}}%
        {\vizbox{\vizfbox}\includegraphics[height=1.85cm,trim=70 50 70 70,clip]{rapps_cave/images/rapps3.eps}}
      }%
      \subfloat[\label{sfig:rapps_depth_camera}]{%
        \ifthenelse{\equal{\arxivmode}{true}}%
        {\includegraphics[height=1.85cm,trim=60 20 40 80,clip]{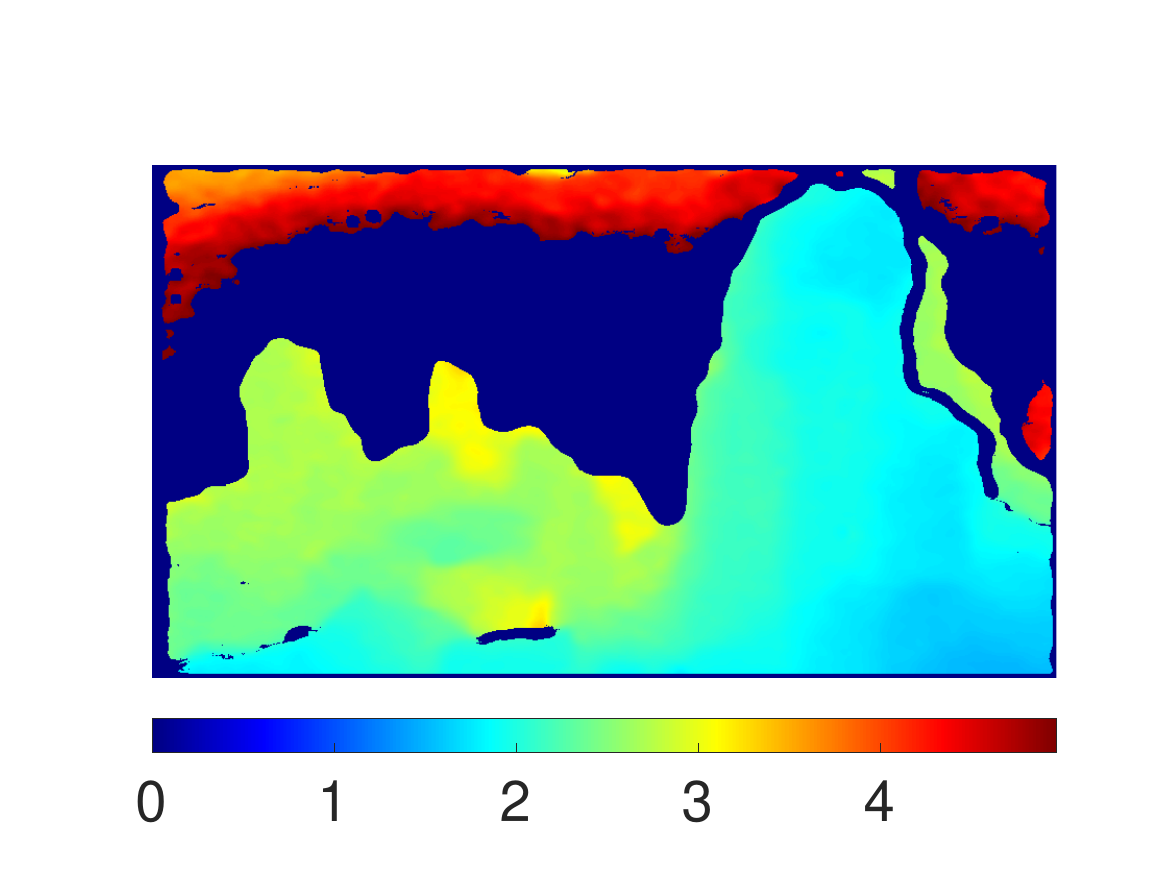}}%
        {\vizbox{\vizfbox}\includegraphics[height=1.85cm,trim=60 20 40 80,clip]{rapps_cave/images/rapps2.eps}}
      }%
      \subfloat[\label{sfig:rapps_resampled_gmm}]{%
        \ifthenelse{\equal{\arxivmode}{true}}%
        {\includegraphics[height=1.85cm,trim=90 80 90 80,clip]{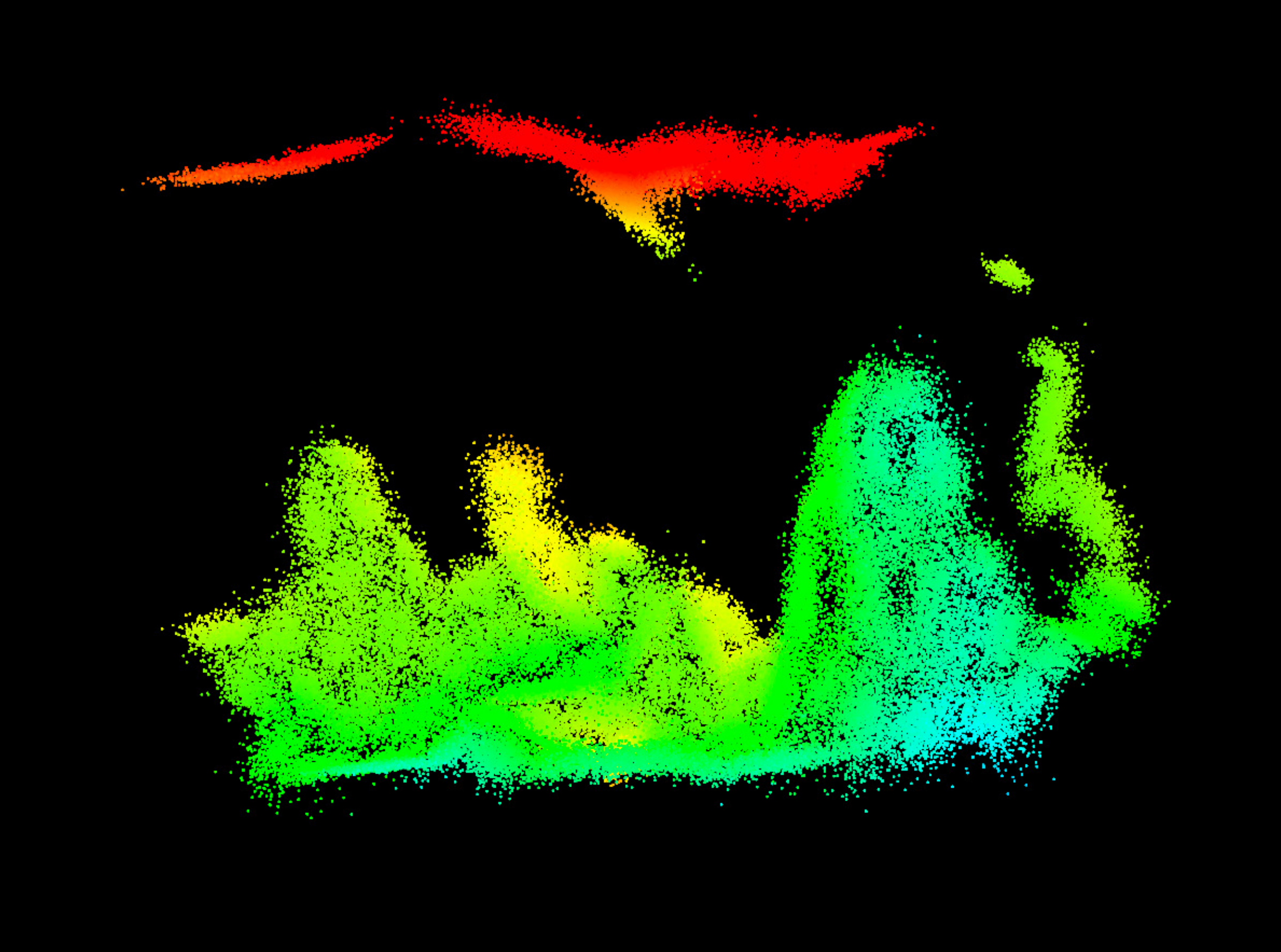}}%
        {\vizbox{\vizfbox}\includegraphics[height=1.85cm,trim=90 80 90 80,clip]{rapps_cave/images/rapps6.eps}}
      }%
  \caption{\label{fig:contrast_mapping} Data and imagery excerpted
from an exploration trial in a cave in West Virginia. \protect\subref{sfig:robot_mapping}
An autonomous aerial system explores and maps a
formation.~\protect\subref{sfig:rapps_color_camera} A color image
taken by the aerial system during exploration (the formation shown
in~\protect\subref{sfig:robot_mapping} can be seen on the left-hand
side of~\protect\subref{sfig:rapps_color_camera}). A depth image taken
at the same time is shown in~\protect\subref{sfig:rapps_depth_camera}
and the resampled GMM map built during flight from the depth image is
shown in~\protect\subref{sfig:rapps_resampled_gmm}. The video of
the exploration trial corresponding
to these images may be found at \url{https://youtu.be/H8MdtJ5VhyU}}
\end{figure}

\IEEEPARstart{T}{he} process of cave surveying, which consists of
marking stations and measuring the distances between them, has changed
relatively little since the 19th
century~\cite[p. 1532]{cave_encyclopedia}. While \citet{cave_encyclopedia}
predicts that advancements in technology may fundamentally change this
method in the 21st century, the substantial advancements in sensing,
3D reconstruction, and autonomy made within the last decade
have not propagated to cave surveying. This paper
addresses this gap in the state of the art through the development and
testing of an autonomous aerial system that explores and maps
caves (see~\cref{fig:contrast_mapping}).

\begin{figure}
  \includegraphics[width=\linewidth]{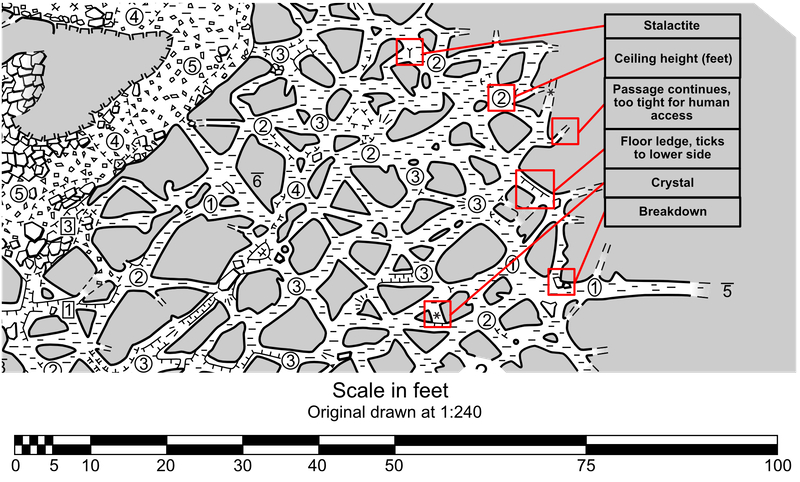}
  \caption{\label{fig:sf_map}An excerpt of the current working map for
    the cave on the Barbara Schomer Cave Preserve in Clarion County,
    PA. The map is encoded with terrain features. Note the passages marked
    too tight for human access in the top-right of the image. Aerial
    robots could be deployed to these areas to collect survey data.
    Image courtesy of B. Ashbrook.}
  \vspace{-0.25cm}
\end{figure}

\begin{figure}
  \includegraphics[width=\linewidth,trim=0 100 0 0,clip]{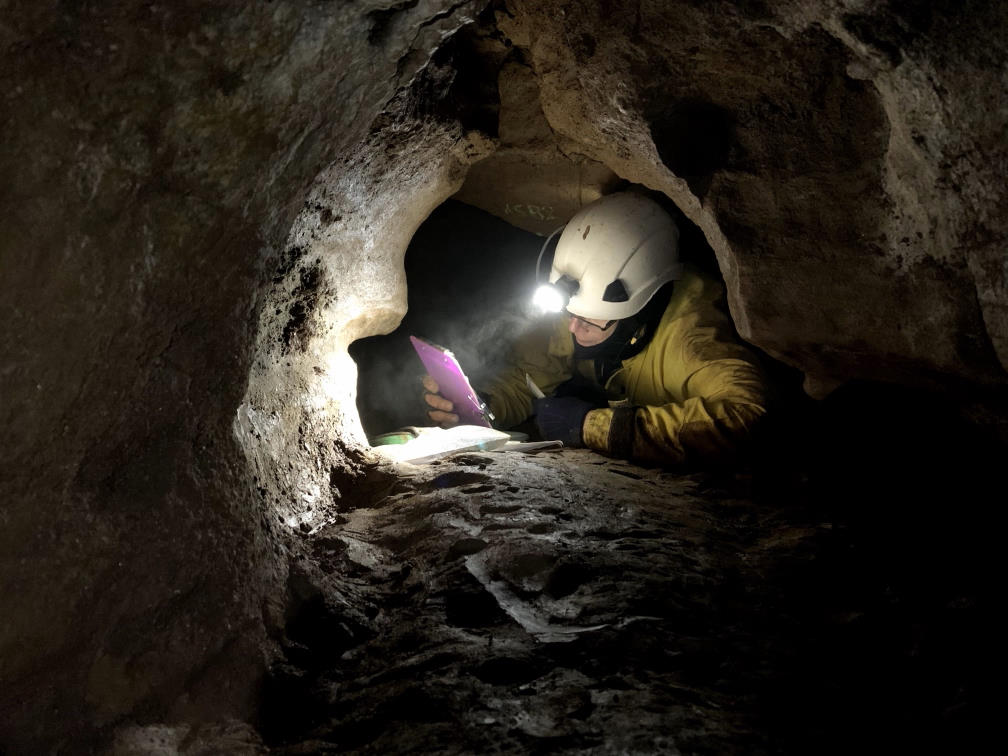}
  \caption{\label{fig:sketching}A caver sketches a passageway to
    produce content for the map shown in~\cref{fig:sf_map}.
    Image courtesy of H. Wodzenski. J. Jahn pictured.}
  \vspace{-0.5cm}
\end{figure}

Traditional cave surveying is challenging because surveyors remain motionless for
extended periods of time and are exposed to water, cool air, and rock,
which can lead to hypothermia~\cite{harler}.  A more serious danger
is getting lost or trapped in a cave~\cite{lostinacave} because
specialized training is required to extract a
physically incapacitated caver.
The Barbara Schomer Cave Preserve in Clarion County, PA, has a cave
that is particularly challenging to survey for two reasons: (1) the
small size of the passages (typical natural passages are
\SI{0.75}{\meter} high and \SI{0.75}{\meter} wide), and (2) the mazelike
nature of the passages that are estimated to be \SI{60}{\kilo\meter}
in length~\cite[p. 10]{nittany_frank}.  50 cave surveyors have been
involved in 30 trips to survey a total of \SI{2522}{\meter} of passage
(B. Ashbrook, personal communication, March 1, 2020).  Each trip is
4-5 hours in duration and there are currently over 90 unexplored
leads in the cave.~\cref{fig:sf_map} illustrates an excerpt of the
current working map of the cave and~\cref{fig:sketching} illustrates
a caver sketching for the cave survey.

Robotic operations in undeveloped subterranean environments are
challenging due to limited or nonexistent communication
infrastructure that increases the risk of failure from data
dropouts.~\citet{kostas2019fsr} find that autonomy is necessary when
operating underground because reliable high-bandwidth wifi connections
are impossible to maintain after making turns.
Low-frequency radio is
commonly used in subterranean environments because it penetrates rock
better than high-frequency radios (e.g.,
walkie-talkies)~\cite{gibson2010channel}, but the disadvantage is that
radio has low-bandwidth and data dropouts increase as range increases.
Given these constraints, cave surveying robots must be autonomous to
robustly operate in the presence of communication dropouts and
must also represent the environment in a way that is both high-resolution yet
compact so that data can be transmitted over low-bandwidth connections
to operators.  The ability to transmit high-resolution maps is
desirable because human operators can direct exploration towards a
particular lead of interest if the robot enters a passageway with
multiple outgoing leads.


GMMs are ideal for perceptual modeling in
the cave survey context because they represent depth information
compactly. This paper seeks to address the problem of robotic
exploration under constrained
communication by proposing an
autonomous system that leverages these compact perceptual models to
transmit high-resolution information about the robot's surroundings
over low-bandwidth connections.  Depth sensor observations are encoded
as GMMs and used to maintain a consistent local occupancy grid map. A
motion planner selects smooth and continuous
trajectories that greedily maximize an information-theoretic objective
function.

This work presents an extended version of prior
work~\cite{Michael-RSS-19} introducing all contributions:
\begin{enumerate}
  \item a method for real-time occupancy reconstruction from GMMs with LiDAR sensor observations,
  \item an information-theoretic exploration system that leverages the occupancy modeling technique, and
  \item evaluation of the exploration system in simulation and
real-world experiments.
\end{enumerate}
This manuscript presents the following additional contributions:
\begin{enumerate}
  \item an extension of the real-time occupancy reconstruction from
GMMs that can accomodate limited field of view depth cameras,
  \item a motion planning framework that performs well for both LiDAR
  and limited field of view depth cameras, and
  \item extensive evaluation of the exploration system in simulation
and on hardware in a show cave \footnote{A cave that has been made
accessible to the public.} in Pennsylvania and a wild
cave\footnote{Wild caves have not been altered to provide access to
  the public so they often present dangers. These dangers can be
  mitigated with special caving gear, knowledge, and experience.} in
West Virginia.
\end{enumerate}
The paper is organized as follows:~\cref{sec:related_work} surveys
related work, \cref{sec:method,,sec:mapping,,ssec:planning} describe
the methodology, \cref{sec:results} presents the experimental
design and results,~\cref{sec:discussion} details the
limitations of the approach, and~\cref{sec:conclusion} concludes with
a discussion of future work.


\section{Related Work\label{sec:related_work}}
State-of-the-art robotic technologies fail to provide rapid
autonomy and persistent situational awareness that meet the
operational challenges of subterranean
environments~\cite{tardioli2019ground}.  The DARPA Subterranean
Challenge~\cite{chung2017darpa} was created to foster the development
of rapid autonomous missions through enhanced situational awareness to
overcome the degraded environmental conditions, severe communication
constraints, and expansive nature that are characteristic of
subterranean regimes. As a result, several works have been
developed pursuant to mine and urban autonomy and exploration.
\Cref{ssec:subt_autonomy} provides an overview of the state of the art in subterranean
autonomy.~\Cref{ssec:cave_mapping} reviews
recent cave mapping works and~\cref{ssec:occ_modeling_exploration}
details works in occupancy modeling for exploration.

\subsection{Exploration in Subterranean Environments\label{ssec:subt_autonomy}}
A significant challenge in cave surveying is
the ability to access leads and passageways that may be
unreachable to humans as shown in~\cref{fig:sf_map}. If the
lead contains a passageway with several outgoing leads, it
is useful for an operator to have the ability to direct exploration in
one direction over others (e.g., one passageway is larger
than another or contains more interesting features).
~\citet{Murphy2009} find that semiautonomy
in the form of autonomous navigation coupled with human-assisted
perception is preferred over total autonomy for robots
operating in subterranean domains given that teleoperation was
required to recover the autonomous \SI{700}{\kilo\gram}
ATV-type Groundhog vehicle when it became stuck in a mine.
However, human-assisted perception is predicated on the ability
to transfer perceptual information to human operators.
\citet{Baker2006} develop Groundhog to collect data via
teleoperation and build a map in postprocessing with the
approach of~\cite{montemerlo2002}.~\citet{groundhog} extend the
capabilities of the Groundhog robot by developing a sense, plan, act
cycle to autonomously map an abandoned mine that takes 90 seconds due
to limited onboard processing.

\citet{ebadi2020lamp} develop an architecture for LiDAR-based
multi-robot SLAM that removes outlier loop closures via Incremental
Consistent Measurement (ICM) Set Maximization and enables input from
multiple sources and a human operator. The approach relies on VLP-16
Puck Lite LiDARs, which are relatively heavy and necessitate using
ground vehicles as compared to the small aerial systems used in this
work. The approach is tested in a mine that has relatively flat
floors as compared to the highly uneven nature of the cave passages
explored in this work. A limitation expressed in the work is the
need for a compressed map representation to overcome the communication
bottlenecks when transmitting visualizations to the user.

\citet{dang2019graph} demonstrate exploration in mine
environments that employs a global planner when the local planner is
unable to identify paths that increase information gain. The aerial
system is large to accomodate the heavy LiDAR sensor which is
appropriate for mine environments built for easy traverse by people
but is too large for the tight crawlways of caves.

\subsection{Cave Mapping\label{ssec:cave_mapping}}
Few works consider methods to map caves with aerial or
ground robotic systems.~\citet{kaul2016continuous} develop the
Bentwing robot to produce maps of cave environments. The approach,
which requires a pilot to remotely operate the vehicle, may be challenging due to the
cognitive load required to stabilize attitude and position
simultaneously. The Bentwing uses a rotating 2D laser scanner to
collect data that is postprocessed into a globally consistent map.
Similarly,~\citet{tabib2019fsr} develop a simultaneous localization
and mapping strategy that represents sensor observations using GMMs
and produces maps via postprocessing from data collected from an aerial
system equipped with a 3D LiDAR.
In contrast, this work produces a map in real-time onboard the
autonomous robot, which is suitable for information-theoretic planning and
sufficiently compact to be transmitted to operators over
low-bandwidth connections for use in human-assisted perception.

The DEep Phreatic THermal eXplorer (DEPTHX) vehicle was developed to
explore and characterize the biology of the Sistema Zacat\'on cenotes,
or underwater sinkholes, in Tamaulipas, Mexico, as an analog mission for the search for life
underneath Europa's ice~\cite{billstone}. The probe combines LiDAR to
map above the water table with sonar to map phreatic zones.  The
authors employ a Deferred Reference Octree data structure to represent
the environment and mitigate the memory required to represent the
underlying evidence grid.  The cost of
transmitting data to enable information sharing with human operators
is not considered.

\subsection{Occupancy Modeling for Exploration\label{ssec:occ_modeling_exploration}}
While many exploration approaches have leveraged voxel-based occupancy
modeling strategies for information-theoretic
planning~\cite{charrow2015icra}, the large memory demands of using
occupancy grid maps remains. Octomap~\cite{hornung2013octomap}
leverages an octree data structure to represent the environment at
multiple resolutions and stores a probability of occupancy
in each cell. A challenge of using octree structures for
dynamically growing maps is that the time complexity of an access or query is $\mathcal{O}(\log n)$~\cite{oleynikova2017voxblox}.
Voxblox~\cite{oleynikova2017voxblox} represents the environment
using a Truncated Signed Distance Field (TSDF) and employs voxel hashing, which yields
$\mathcal{O}(1)$ look-up and insertion times, but suffers from the same
discretization challenges as the occupancy grid map.
The Normal Distribution Transform
Occupancy Map (NDT-OM) attempts to mitigate the large memory demands of occupancy grid maps~\cite{saarinen2013normal} by encoding a Gaussian density into
occupied voxels based on the heuristic that
larger voxels are sufficient to represent the environment at the
same fidelity as a traditional occupancy grid map. However,
this technique suffers from the same drawbacks as the occupancy
grid map for exploration in large environments.

To overcome this limitation, this work builds upon prior work
by~\citet{omeadhra2018variable} that compactly represents sensor observations
as GMMs for the purpose of occupancy reconstruction by developing
real-time local occupancy mapping for information-theoretic planning
using both \SI{360}{\degree} and limited FoV sensor models. Because
sensor observations are stored as GMMs, local occupancy maps are
constructed as needed and only GMMs need to be transmitted between
the robot and operator, which results in a much smaller memory
footprint.  Another advantage is that occupancy grid maps of variable
resolution could be used to plan paths at various resolutions (though
this is not demonstrated in this work).

While GMMs have been used for compact perceptual
modeling~\cite{srivastava2018efficient}, occupancy
modeling~\cite{omeadhra2018variable}, and multi-robot
exploration~\cite{corah2018ral}, these works do not demonstrate
real-time operation. Our prior work~\cite{Michael-RSS-19} addresses
this gap in the state of the art by proposing an exploration system
that leverages GMMs for real-time 3D information theoretic planning
and perceptual modeling using a \SI{360}{\degree} field of view (FoV)
sensor on computationally constrained platforms.  The choice of
sensor field of view (FoV) has implications for
mapping, planning, and hardware design so this paper extends the prior
work by considering both \SI{360}{\degree} and limited FoV depth
sensors.


\section{Overview\label{sec:method}}
\begin{figure}
  \centering
  \vizbox{\vizfbox}{\includegraphics[width=\linewidth,trim=0 0 0 0]{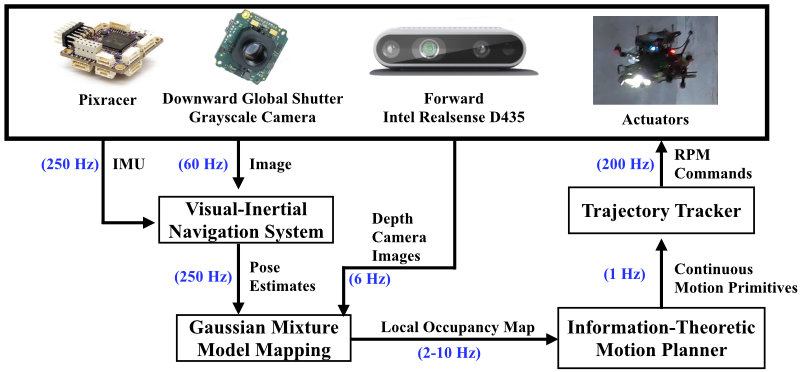}}
  \caption{\label{fig:methodology_overview}Overview of the autonomous exploration system
    presented in this work. Using pose estimates from a visual-inertial navigation system
    (\cref{ssec:vins}) and depth camera observations, the mapping method
    (\cref{ssec:preliminaries}
    and~\cref{ssec:local_grid_map}) builds a memory-efficient approximate continuous belief
    representation of the environment while creating local occupancy grid maps in real-time.
    A motion primitives-based information-theoretic planner (\cref{ssec:planning}) uses this
    local occupancy map to generate snap-continuous forward-arc motion primitive trajectories
  that maximize the information gain over time.}
\end{figure}

The exploration system consists of mapping,
information-theoretic planning, and a monocular visual-inertial
navigation system (Fig.~\ref{fig:methodology_overview}). A brief
review of GMMs is detailed
in~\cref{ssec:preliminaries}.~\Cref{ssec:local_grid_map} develops the
GMM-based local occupancy grid mapping strategy used by the planning
approach to generate continuous trajectories that maximize an
information-theoretic objective (\cref{ssec:planning}).


\section{Mapping}\label{sec:mapping}
\subsection{Gaussian Mixture Models for Perception\label{ssec:preliminaries}}
\begin{figure*}
  \centering
  \subfloat[Color Image \label{sfig:color_image}]{\vizbox{\vizfbox}{\includegraphics[width=0.144\linewidth,trim=40 12 20 0]{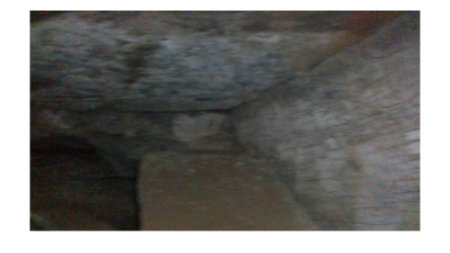}}}%
  \subfloat[Depth Image \label{sfig:depth_image}]{\vizbox{\vizfbox}{\includegraphics[width=0.17\linewidth,trim=32 67 32 65,clip]{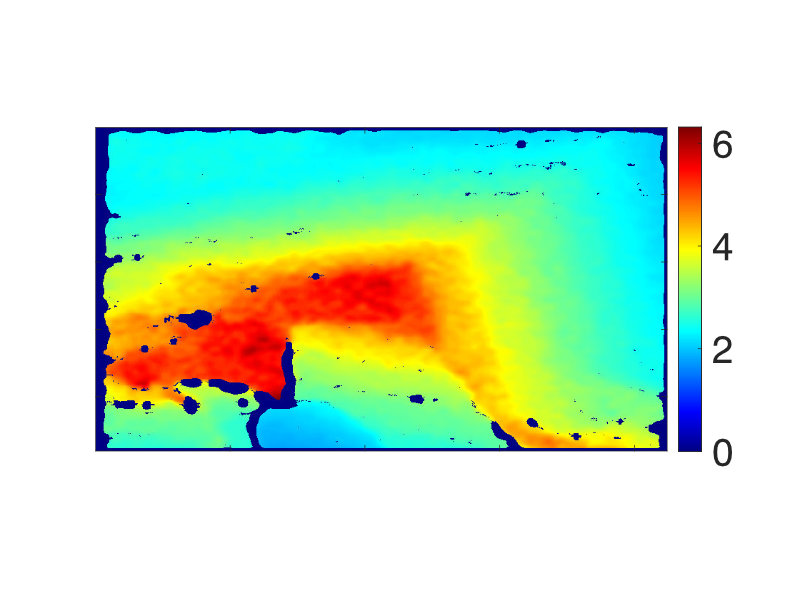}}}%
  \subfloat[PointCloud \label{sfig:cave_pointcloud}]{\vizbox{\vizfbox}{\includegraphics[width=0.165\linewidth,trim=0 40 0 0,clip]{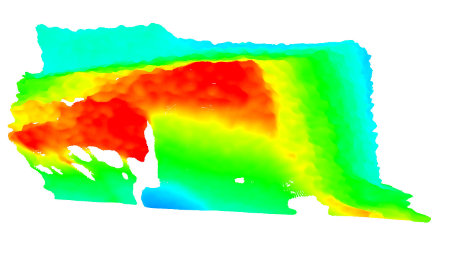}}}
  \subfloat[Free Space Windows \label{sfig:cave_occ_free}]{\vizbox{\vizfbox}{\includegraphics[width=0.165\linewidth,trim=0 40 0 0,clip]{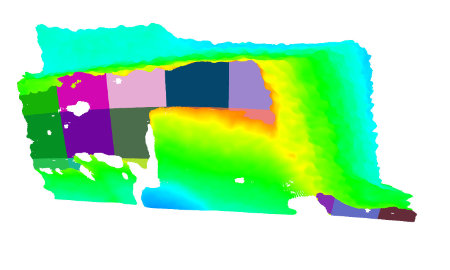}}}%
  \subfloat[GMM \label{sfig:cave_gmms}]{\vizbox{\vizfbox}{\includegraphics[width=0.165\linewidth,trim=0 40 0 0,clip]{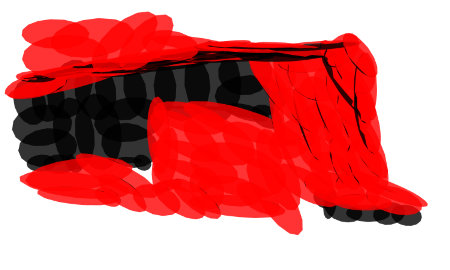}}}%
  \subfloat[Resampled Data\label{sfig:cave_gmms_resampled}]{\vizbox{\vizfbox}{\includegraphics[width=0.165\linewidth,trim=0 40 0 0,clip]{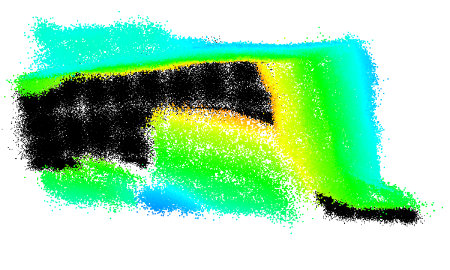}}}%
  \caption{\label{fig:windowing_strategy}
    Overview of the approach to transform a sensor observation into
free and occupied GMMs.
    \protect\subref{sfig:color_image} A color image taken onboard
the robot exploring Laurel Caverns.
    \protect\subref{sfig:depth_image} A depth image corresponding to
the same view as the color image with distance shown as a heatmap on
the right hand side (in meters).
    \protect\subref{sfig:cave_pointcloud} illustrates the point cloud
representation of the depth image.
    \protect\subref{sfig:cave_occ_free} In the mapping approach,
points at a distance smaller than a user-specified max range $r_d$ (in
this case $r_d = \SI{5}{\meter}$) are considered to be occupied, and a
GMM is learned using the approach detailed
in~\cref{sssec:occupied_space}.  Points at a distance further than
$r_d$ are considered free, normalized to a unit vector, and projected
to $r_d$. The free space points are projected to image space and
windowed using the technique detailed in~\cref{sssec:free_space} to
decrease computation time. Each window is shown in a different color.
    \protect\subref{sfig:cave_gmms} The GMM representing the
occupied-space points is shown in red and the GMM representing the
free space points is shown in black.  Sampling $2\times10^5$ points
from the distribution yields the result shown
in~\protect\subref{sfig:cave_gmms_resampled}. The number of points to
resample is selected for illustration purposes and to highlight that
the resampling process yields a map reconstruction with an arbitrary
number of points.}
\vspace{-0.5cm}
\end{figure*}

The approach leverages GMMs to compactly encode sensor observations for
transmission over low-bandwidth communications channels. The GMM provides a
generative model of sensor observations from which occupancy may be
reconstructed by resampling from the distribution and raytracing through a local
occupancy grid map. Formally, the GMM is a weighted sum of $M$
Gaussian probability density functions (PDFs). The probability density of the
GMM is expressed as
\begin{align*}
  p(\bbf{x}|\gmmparams) &= \sum\limits_{m=1}^M \pi_m \mathcal{N}(\bbf{x} | \mean_m, \cov_m)
\end{align*}
where $p(\bbf{x}|\gmmparams)$ is the probability density for the
D-dimensional random variable $\bbf{x}$ and is parameterized by
$\gmmparams = \{\pi_m, \mean_m, \cov_m \}_{m=1}^M$. $\pi_m \in
\mathbb{R}$ is a weight such that $\sum_{m=1}^M \pi_m = 1$
and $0 \leq \pi_m \leq 1$, $\mean_m$ is a mean, and $\cov_m$ is a
covariance matrix for the $m^{th}$ $D$-dimensional Gaussian
probability density function of the distribution. The multivariate
probability density for $\bbf{x}$ is written as
\begin{align*}
  \mathcal{N}(\bbf{x} | \mean_i,\cov_i) &=
                                          \frac{|\cov_i|^{-1/2}}{(2\pi)^{D/2}}\exp\Big(
                                          -\frac{1}{2}(\bbf{x} - \mean_i)^T\cov_i^{-1}(\bbf{x}-\mean_i)\Big).
\end{align*}
In this work, a depth
observation taken at time $t$ and consisting of $N$ points,
$\depthobs_t = \{\point_t^1, \ldots, \point_t^n, \ldots,
\point_t^N\}$, is used to learn a GMM.
Estimating optimal GMM parameters $\gmmparams$ remains an open area of
research~\cite{hosseini2017alternative}. This work utilizes the
Expectation Maximization (EM) algorithm to solve the
maximum-likelihood parameter estimation problem, which is guaranteed
to find a local maximum of the log likelihood
function~\cite{bishop2007pattern}. To make the optimization tractable, EM introduces latent
variables $\latentvarmat = \{\latentvar_{nm}\}$ for each point
$\point_t^n$ and cluster $m$ and iteratively performs two
steps: expectation (E) and maximization (M)~\cite{bishop2007pattern,bilmes1998gentle,eckart2015mlmd}.

The E step calculates the expected value of the complete-data
log-likelihood $\ln p(\depthobs_t, \latentvarmat | \gmmparams)$ with respect to
the unknown variables $\latentvarmat$ given the observed data $\depthobs_t$ and
current parameter estimates $\gmmparams^i$, which is written as $E[\ln p
(\depthobs_t, \latentvarmat| \gmmparams)|\depthobs_t,
\gmmparams^i]$~\cite{bilmes1998gentle}. This amounts to evaluating the posterior
probability, $\responsiblityval_{nm}$, using the current parameter values
$\gmmparams^i$ (shown in ~\cref{eq:estep})~\cite{bishop2007pattern}
\begin{align}
\responsiblityval_{nm} = \frac{\pi_m \mathcal{N}(\point_{t}^{n} | \mean_m^i, \cov_m^i)}{\sum\limits_{j=1}^M \pi_j
                \mathcal{N}(\point_{t}^{n} | \mean_j^i,
                                \cov_j^i)}, \label{eq:estep}
\end{align}
where $\responsiblityval_{nm}$ denotes the responsibility that
  component $m$ takes for point $\point_t^n$.
The M step maximizes the expected log-likelihood using the current
  responsibilities,
  $\responsiblityval_{nm}$, to obtain updated parameters, $\gmmparams^{i+1}$ via
the following:
\begin{align}
  \mean_m^{i+1} &= \sum_{n=1}^{N} \frac{\beta_{nm}\point_{t}^{n}}{\sum_{n=1}^N \beta_{nm}}\\
  \cov_m^{i+1} &= \sum\limits_{n=1}^{N}
                            \frac{\beta_{nm}(\point_{t}^{n}
                            - \mean_m^{i+1}) (\point_{t}^{n} -
                            \mean_m^{i+1})^T}{\sum_{n=1}^N \beta_{nm}}\\
  \pi_m^{i+1} &= \frac{\sum_{n=1}^N \beta_{nm} \bbf{x}_n}{\sum_{n=1}^N
                \beta_{nm}}.
\end{align}
Every iteration of EM is guaranteed to increase the log likelihood and iterations
are performed until a local maximum of the log likelihood is achieved~\cite{bishop2007pattern}.

The E step is computationally expensive because a responsibility
$\responsiblityval_{nm}$ is calculated for each cluster $m$ and point
$\point_t^n$, which amounts to $NM$ responsibility calculations. In
the M step, every parameter must be updated by iterating over all $N$
samples in the dataset. In practice, a responsibility matrix
$\responsiblitymat \in \mathbb{R}^{N\times M}$ is maintained whose
entries consist of the $\responsiblityval_{nm}$ to estimate the
parameters $\gmmparams$.

Following the work of~\citet{omeadhra2018variable}, distinct
occupied $\mathcal{G}(\bbf{x})$ (detailed in \cref{sssec:occupied_space}) and
free $\mathcal{F}(\bbf{x})$ (detailed in \cref{sssec:free_space}) GMMs
are learned to compactly represent the density of points observed in
the environment (\cref{fig:windowing_strategy}).
The process by which $\mathcal{F}(\bbf{x})$ and
$\mathcal{G}(\bbf{x})$ are created is illustrated
in~\cref{sfig:cave_pointcloud,sfig:cave_occ_free}.  Because the GMM is
a generative model, one may sample from the distribution
(\cref{sfig:cave_gmms_resampled}) to generate points associated with
the surface model and reconstruct occupancy (detailed in \cref{ssec:local_grid_map}).

\subsubsection{Occupied Space\label{sssec:occupied_space}}
For points with norms less than a user-specified maximum range $r_d$,
the EM approach is adapted from~\cite{tabib2019fsr} to accept points
that lie within a Mahalanobis distance of $\lambda$.
Because Gaussians fall off quickly, points far away from a
given density will have a small effect on the updated parameters for
that density.  By reducing the number of points, this decreases the
computational cost of the EM calculation. Only points that have a
value smaller than $\lambda$ are considered (i.e., points larger
than $\lambda$ are discarded):
\begin{align}
  \label{eq:mahalanobis_approx}
  \lambda > \sqrt{(\bbf{x}_n - \bbf{\mu}_m^1)^T (\bbf{\Lambda_m}^1)^{-1}(\bbf{x}_n - \bbf{\mu}_m^1)}
\end{align}
where the superscript 1 denotes the initialized values for the mean,
covariance, and weight.  This approach differs from our prior
work~\citet{Michael-RSS-19}; we utilize the approach in~\cite{tabib2019fsr}
as it yields greater frame-to-frame registration accuracy in practice.
Frame-to-frame registration is not
used in this work and is left as future work.

\subsubsection{Free Space\label{sssec:free_space}}
To learn a free space distribution, points with norms that exceed
the maximum range $r_d$ are projected to
$r_d$.  The EM approach from~\cref{sssec:occupied_space} is used to
decrease the computational cost of learning the distribution. To
further reduce cost, the free space points are split into
windows in image space and GMMs consisting of $n_f$ components are
learned for each window.
The windowing strategy is employed for learning distributions over
free space points because it yields faster results and the
distributions cannot be used for frame-to-frame registration.
The number of windows and components per window is selected empirically.
~\cref{sfig:cave_occ_free} illustrates the
effect of the windowing using colored patches
and~\cref{sfig:cave_gmms} illustrates the result of this windowing
technique with black densities. Once the free space distributions
are learned for each window the windowed distributions are
merged into a single distribution.

Let $\mathcal{G}_i(\bbf{x})$ be a GMM trained
from $N_i$ points in window $i$ and let $\mathcal{G}_j(\bbf{x})$ be a GMM
trained from $N_j$ points in window $j$, where $\sum_{w=1}^W N_w = N$ for
sensor observation $\depthobs_t$ and $W$ windows. $\mathcal{G}_j(\bbf{x}) = \sum_{k=1}^K
\tau_k \mathcal{N}(\bbf{x} | \bbf{\nu}_k, \bbf{\Omega}_k)$ may be merged into
$\mathcal{G}_i(\bbf{x}) = \sum_{m=1}^M \pi_m \mathcal{N}(\bbf{x} |
\bbf{\mu}_m, \bbf{\Lambda}_m)$ by concatenating the means, covariances, and
weights. However, care must be taken when merging the weights as they must be
renormalized to sum to 1~\cite{shobhit_thesis}. The weights are renormalized
via~\cref{eq:support_size,eq:weight_update}:
\begin{align}
  N^* &= N_i + N_j \label{eq:support_size}\\
  \bbf{\pi}^* &= \begin{bmatrix}
    \frac{N_i\pi_1}{N^*} &  \ldots & \frac{N_i\pi_m}{N^*} &
    \frac{N_j\tau_1}{N^*} & \ldots & \frac{N_j\tau_k}{N^*}
  \end{bmatrix}^T \label{eq:weight_update}
\end{align}
where $m \in [1,\ldots,M]$ and $k \in [1,\ldots,K]$ denote the mixture
component in GMMs $\mathcal{G}_i(\bbf{x})$ and $\mathcal{G}_j(\bbf{x})$,
respectively. $N^* \in \mathbb{R}$ is the sum of the support sizes of
$\mathcal{G}_i(\bbf{x})$ and $\mathcal{G}_j(\bbf{x})$. $\bbf{\pi}^* \in
\mathbb{R}^{M+K}$ are the renormalized weights. The means and covariances are
merged by concatenation.

\begin{figure*}
  \centering
  \subfloat[\label{sfig:bbxa}]{\includegraphics[height=4cm,trim=40 0 75 10,clip]{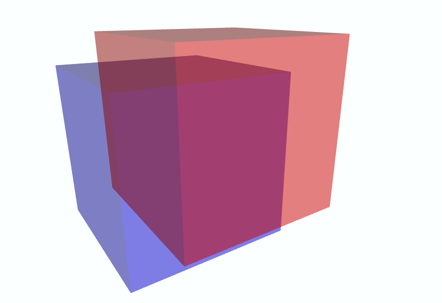}}%
  \subfloat[\label{sfig:bbxb}]{\includegraphics[height=4cm,trim=40 0 75 10,clip]{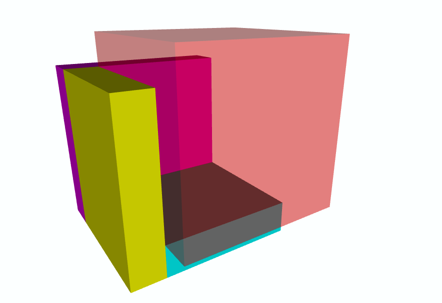}}%
  \subfloat[\label{sfig:novelbbx1}]{\includegraphics[height=4cm,trim=100 0 100 0,clip]{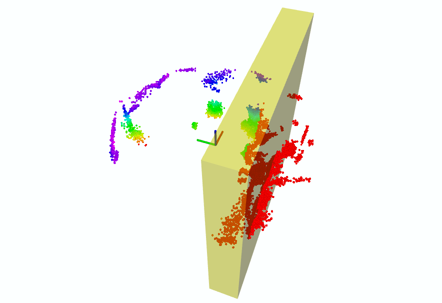}}%
  \subfloat[\label{sfig:novelbbx2}]{\includegraphics[height=4cm, trim=100 0 100 0,clip]{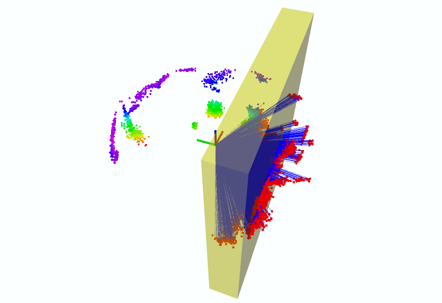}}
  \caption{Overview of the method by which occupancy is reconstructed.
    \label{fig:bbx}\protect\subref{sfig:bbxa} The blue bounding box $\bbx_{t+1}$
is centered around $\robotpose_{t+1}$ and red bounding box $\bbx_t$ is centered
at $\robotpose_{t}$.~\protect\subref{sfig:bbxb} illustrates the novel bounding
boxes in solid magenta, teal, and yellow colors that represent the set
difference $b_{t+1} \setminus b_{t}$.~\protect\subref{sfig:novelbbx1}
Given a sensor origin shown as a triad, resampled pointcloud,
and novel bounding box shown in yellow, each
ray from an endpoint to the sensor origin is tested to determine if an intersection
with the bounding box occurs. The endpoints of rays that intersect the bounding box are
shown in red.~\protect\subref{sfig:novelbbx2} illustrates how the bounding box
occupancy values are updated. Endpoints inside the yellow volume update cells
with an occupied value. All other cells along the ray (shown in blue)
are updated to be free.}
\end{figure*}

\subsection{Local Occupancy Grid Map\label{ssec:local_grid_map}}
The occupancy grid map~\cite{thrun2005probabilistic} is a probabilistic
representation that discretizes 3D space into finitely many grid cells
$\mathbf{m} = \{m_1, ..., m_{|\mathbf{m}|}\}$. Each cell is assumed to be
independent and the probability of occupancy for an individual cell is denoted
as $p(m_i | \mathbf{X}_{1:t}, \depthobs_{1:t})$, where $\mathbf{X}_{1:t}$ represents
all vehicle states up to and including time $t$ and $\depthobs_{1:t}$ represents
the corresponding observations. Unobserved grid cells are assigned the uniform
prior of 0.5 and the occupancy value of the grid cell $m_i$ at time $t$ is
expressed using log odds notation for numerical stability.
\begin{align*}
  l_{t,i} \triangleq \log
  \Bigg(\frac{p(m_i|\depthobs_{1:t},\mathbf{X}_{1:t})}{1-p(m_i|\depthobs_{1:t},
  \mathbf{X}_{1:t})} \Bigg) - l_0
\end{align*}
When a new measurement $\depthobs_t$ is obtained, the occupancy value of cell $m_i$ is
updated as
\begin{align*}
  l_{t,i} \triangleq l_{t-1,i} + L(m_i | \depthobs_t)
\end{align*}
where $L(m_i|\depthobs_t)$ denotes the inverse sensor model of the robot
and $l_0$ is the prior of occupancy~\cite{thrun2005probabilistic}.

Instead of storing the occupancy grid map $\mathbf{m}$ that represents
occupancy for the entire environment viewed since the start of exploration
onboard the vehicle, a local occupancy grid map $\bar{\mathbf{m}}_t$ is
maintained centered around the
robot's pose $\mathbf{X}_t$. The local occupancy grid map moves with the
robot, so when regions of the environment are revisited, occupancy must be
reconstructed from the surface models $\mathcal{G}(\bbf{x})$ and
$\mathcal{F}(\bbf{x})$. To reconstruct occupancy at time $t+1$ given
$\bar{\mathbf{m}}_t$,
the set difference of the bounding boxes $b_t$ and $b_{t+1}$ for
$\bar{\mathbf{m}}_t$ and $\bar{\mathbf{\mathbf{m}}}_{t+1}$, respectively, are used to
compute at most three non-overlapping bounding boxes
(see~\cref{sfig:bbxa,sfig:bbxb} for example). The
intersection of the bounding boxes remains up-to-date, but the occupancy of the
novel bounding boxes must be reconstructed using the surface models
$\mathcal{G}(\bbf{x})$ and $\mathcal{F}(\bbf{x})$. Raytracing is an expensive
operation~\cite{amanatides1987fast}, so time is saved by removing voxels at the
intersection of $b_t$ and $b_{t+1}$ from consideration.
\begin{figure}
  \begin{minipage}[t]{0.6\linewidth}%
    \vspace*{0pt}%
    \ifthenelse{\equal{\arxivmode}{true}}%
      {\includegraphics[]{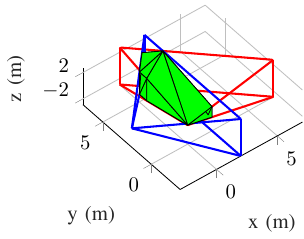}}%
      {\vizbox{\vizfbox}{\trimbox{0.0cm 0cm 0cm 0cm}{\input{./images/intersections.tex}}}}\\
  \end{minipage}%
  \hfill%
  \begin{minipage}[t]{0.37\linewidth}%
    \vspace*{0pt}%
    \caption{\label{fig:polyhedra} For limited FoV sensors, the FoV is approximated by the
      illustrated blue and red rectangular pyramids. These FoVs
      may also be represented as tetrahedra. To determine if a sensor
      position should be stored, the overlapping volume between the
    two approximated sensor FoVs is found.}%
\end{minipage}%
\end{figure}

The local occupancy grid map at time $t+1$, $\bar{\mathbf{m}}_{t+1}$,
is initialized by copying the voxels in local grid
$\bar{\mathbf{m}}_{t}$ at the intersection of $b_{t+1}$ and
$b_{t}$. In practice, the time to copy the local occupancy grid map is
very low (on the order of a few tens of milliseconds) as compared to
the cost of raytracing through the grid. Not all Gaussian densities
will affect the occupancy reconstruction so to identify the GMM
components that intersect the bounding boxes a
KDTree~\cite{blanco2014nanoflann} stores the means of the densities.
A radius equal to twice the sensor's max range is used to identify the
components that could affect the occupancy value of the cells in the
bounding box. A ray-bounding box intersection
algorithm~\cite{williams2005efficient} checks for intersections
between the bounding box and the ray from the sensor origin to the density
mean. Densities that intersect the bounding box are extracted into
local submaps $\bar{\mathcal{G}}(\bbf{x})$ and
$\bar{\mathcal{F}}(\bbf{x})$. Points are sampled from each
distribution and raytraced to their corresponding sensor origin to
update the local grid map (example shown
in~\cref{sfig:novelbbx1,sfig:novelbbx2}).

As the number of mixture components increases over
time in one region, updating the occupancy becomes increasingly
expensive as the number of points needed to resample and raytrace
increases.  The next sections detail how to limit the
potentially unbounded number of points depending on the employed sensor
model.

\subsection{$\SI{360}{\degree}$ FoV Sensor Model}
A small, fixed-size bounding box around the current pose with
half-lengths $h_x$, $h_y$, and $h_z$ is used to determine if a prior
observation was made within the confines of the bounding box. This
bounding box approach works for sensors that have a 360$^{\circ}$
field of view such as the 3D LiDAR used in this work, but does not
readily extend to depth sensors with smaller fields of view (discussed
in the next section). If a
prior observation was made within the bounding box of the current
observation, $\depthobs_t$ is not stored as a GMM. This has the
effect of limiting the number of components that are stored over time.

\subsection{Limited FoV Sensor Model}
The limited FoV sensor model is directional, so it is approximated by two
non-intersecting tetrahedra such that their union forms a rectangular
pyramid (shown in~\cref{fig:polyhedra}). For two sensor FoVs the
intersection between the four pairs
of tetrahedra is calculated and the intersection points found.  The
convex hull of the intersection points is converted to a
polyhedron mesh with triangular facets. The convex hull volume
is found by summing individual volumes of the tetrahedra that make
up the polyhedron~\cite{wolframalpha}. The overlap is estimated as
a percentage of overlapping volume between the two sensor FoVs and a
sensor observation is only stored if its overlap exceeds a
user-defined threshold. In this way, the number of components that represent a given
location can be reduced while ensuring that the environment is covered.


\section{Planning for Exploration\label{ssec:planning}}
\begin{figure}[t]
  \centering
  \subfloat[\label{subfig:single_mpl}]{\includegraphics[width=0.4\linewidth]{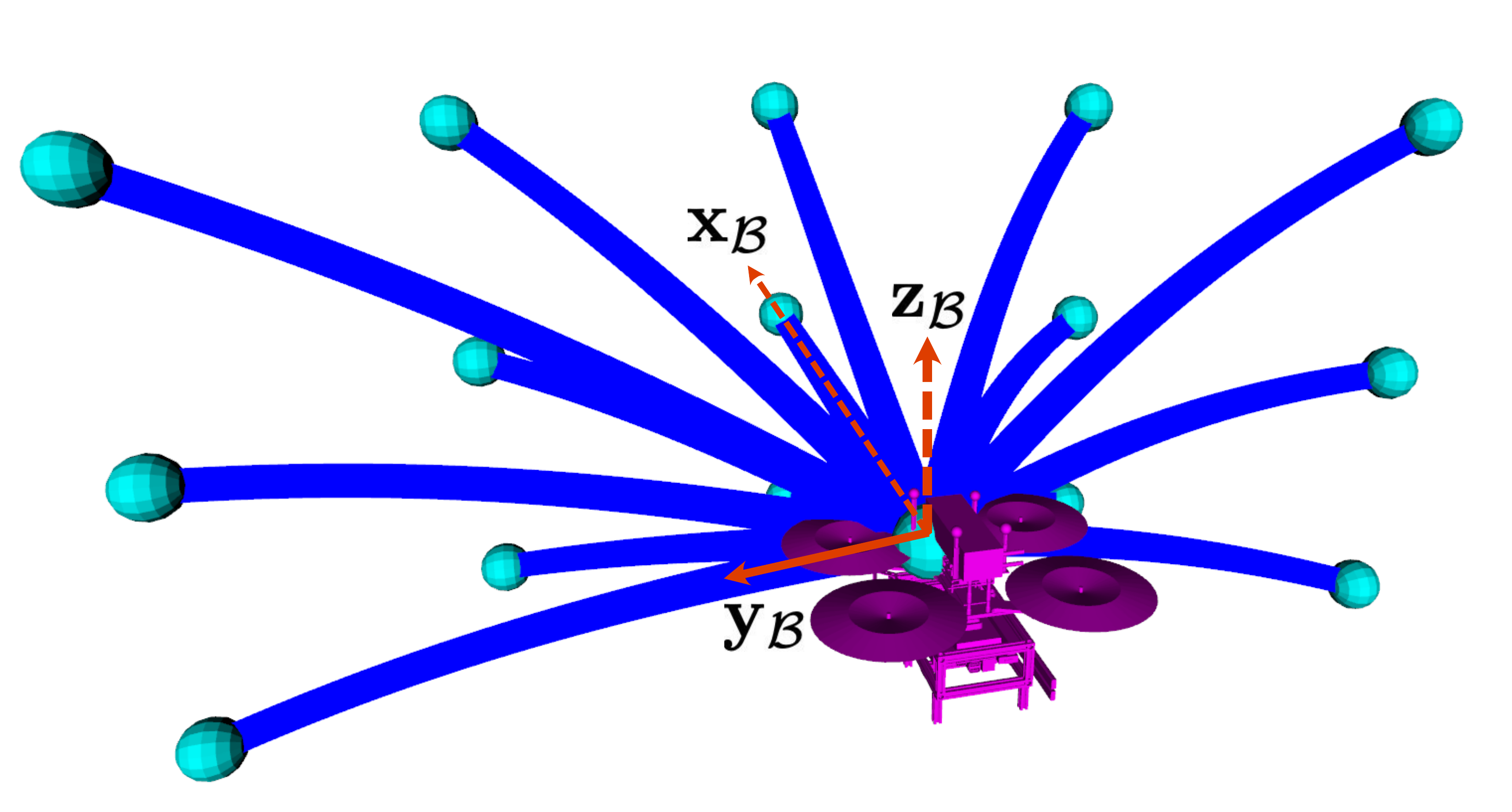}}
  \subfloat[\label{subfig:prims_lidar}]{\includegraphics[width=0.2\linewidth]{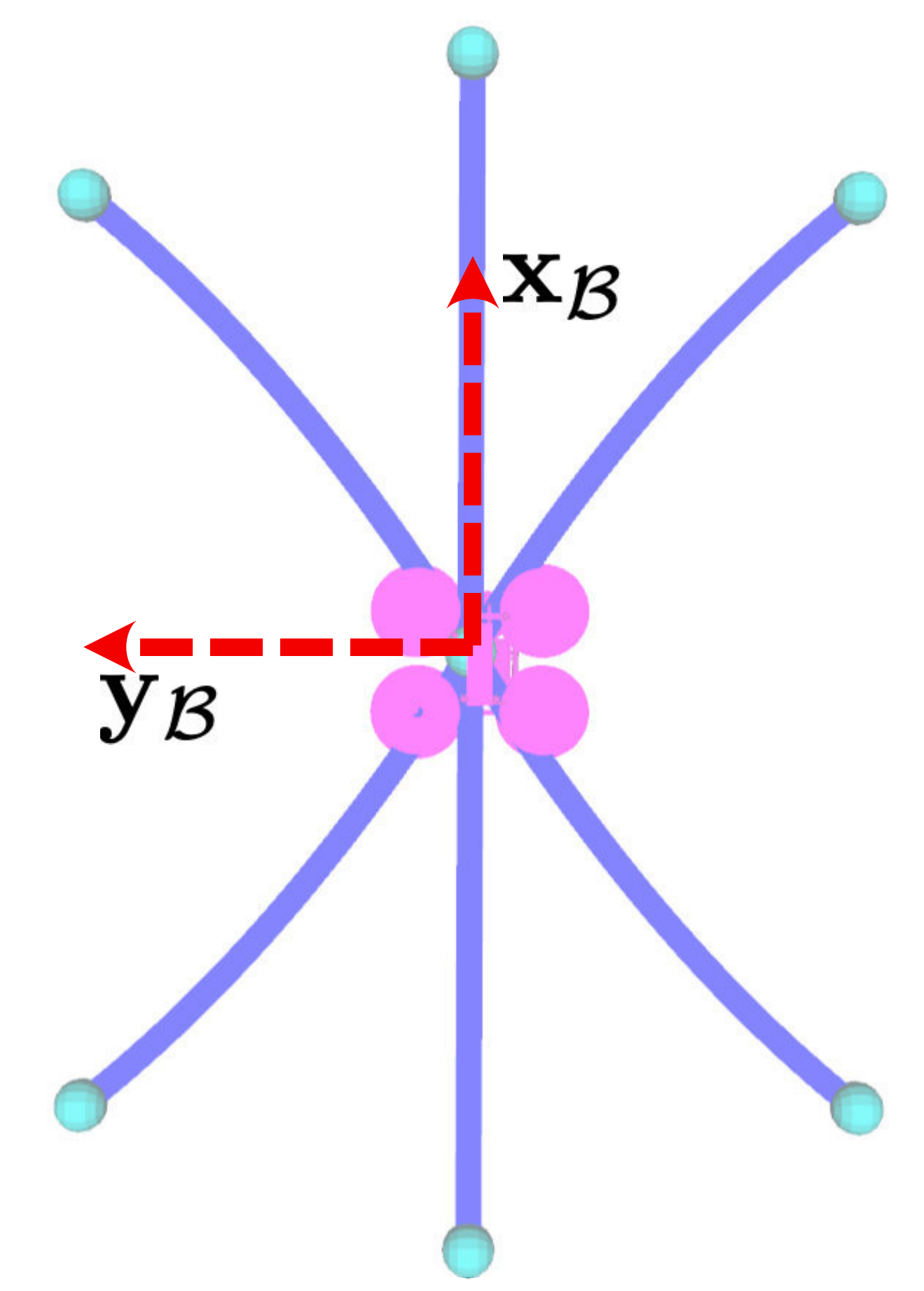}}
  \subfloat[\label{subfig:prims_tof}]{\includegraphics[width=0.4\linewidth]{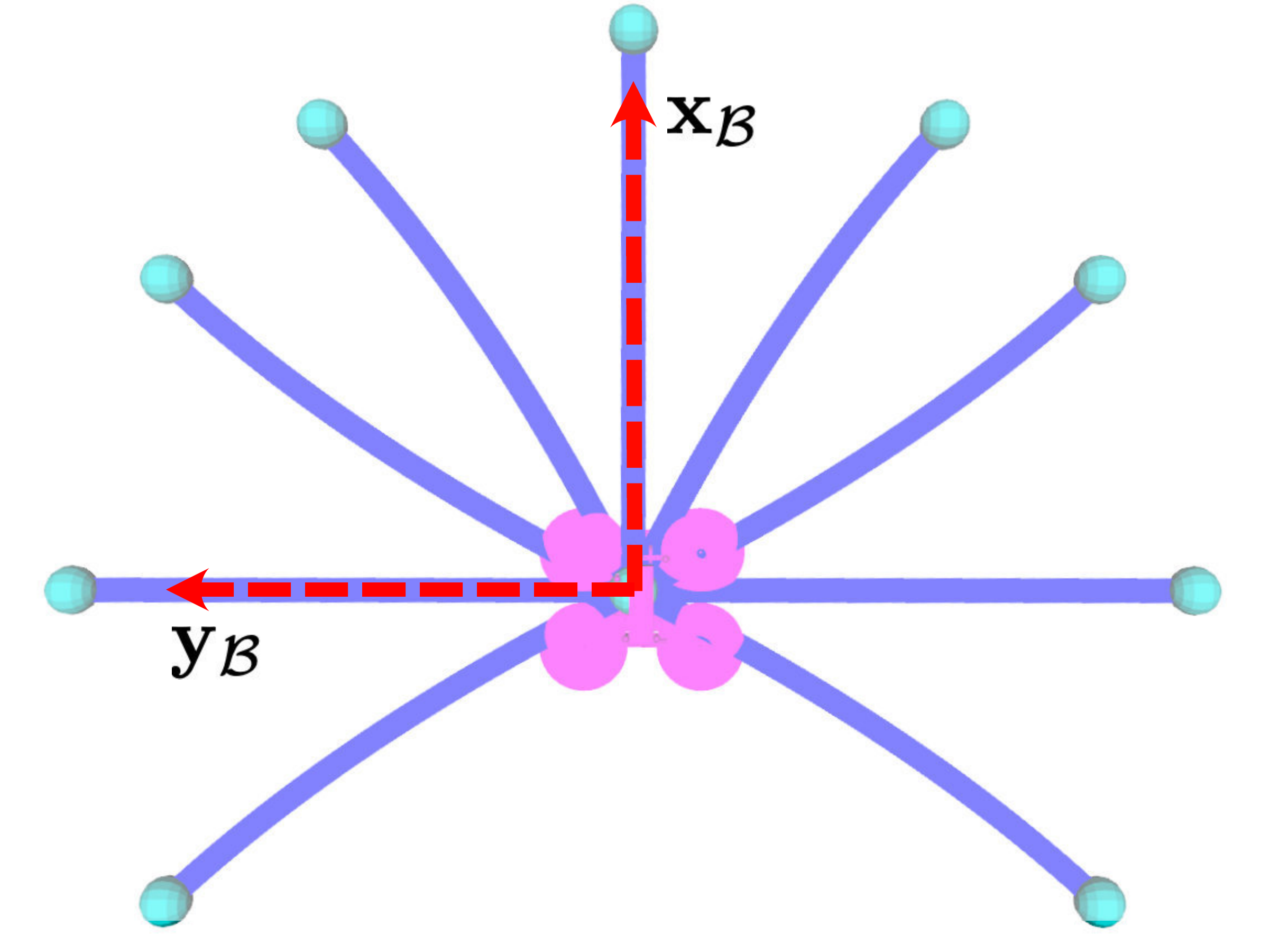}}\\
  \caption{\label{fig:action_generation}Action space design for the proposed
    information-theoretic planner.  \protect\subref{subfig:single_mpl} shows a
    single motion primitive library generated using bounds on the linear
    velocity along $\{\x_\body, \z_\body\}$ and the angular velocity along $\{
    \z_\body \}$.  \protect\subref{subfig:prims_lidar} and
    \protect\subref{subfig:prims_tof} show top-down views of the motion
    primitive library collections used when the sensor model is a
    LiDAR~\citep{Michael-RSS-19} and a depth camera~\citep{goel2019fsr}
    respectively (off-plane primitives are not shown). The proposed planner can
    be used with either of these sensors using the appropriate action space
  designs explained in~\cref{sssec:action_generation}.  }
\end{figure}

A motion planner designed for exploration of \textit{a priori} unknown and
unstructured spaces with an aerial robot must satisfy three requirements: (1)
reduce entropy of the unknown map, (2) maintain collision-free operation, and
(3) return motion plans in real-time. Several previous works provide
information-theoretic frameworks towards meeting these objectives
\citep{julian2014, charrow2015icra, Zhang2019}. \citet{julian2014} use the
Shannon mutual information between the map and potential sensor observations as
a reward function to generate motion plans for exploration; however,
computational requirements limit the number of potential trajectories over
which the reward can be calculated. In contrast, the Cauchy-Schwarz Quadratic
mutual information (CSQMI) has been demonstrated for real-time exploration with
aerial robots~\citep{charrow2015icra, Michael-RSS-19}.  This work utilizes an
information-theoretic planning strategy using CSQMI as the primary reward
function, extending our prior work~\citep{Michael-RSS-19} to support limited
FoV sensors in addition to $360\si{\degree}$ FoV sensors. The proposed
framework can be divided into two stages: (1) action space generation and (2)
action selection.  At the start of any planning iteration, the planner uses the
action generation strategy (detailed in~\cref{sssec:action_generation}) to
generate a set of candidate actions up to a user-specified planning horizon
using motion primitives. The action selector evaluates the collision-free and
dynamically feasible subset of the action space using CSQMI as a reward
function, returning the most informative plan to execute during the next
planning iteration
(see~\cref{sssec:information_theory,,sssec:action_selection}).

\subsection{Action Space Generation}\label{sssec:action_generation}
This section describes: (1) background on the trajectory representation using
motion primitive generation~\citep{yang2017framework}, and (2) the design of the
action space.

\subsubsection{Forward-Arc Motion Primitive} Accurate position control of
multirotors presumes continuity in the supplied tracking references up to
high-order derivatives of position~\citep{Mellinger2012}. To represent a
candidate trajectory, this paper utilizes sequential forward-arc motion
primitives~\citep{yang2017framework}.  We use an extension to this work that
ensures differentiability up to jerk and continuity up to
snap~\citep{spitzer2018}. Given the multirotor state at a time $t$, $\state_t
= [x, y, z, \theta]^\top$, linear velocities in the body frame $[
v_{\x_\body}, v_{\z_\body}]$, and the angular velocity about $\z_\body$ axis,
$\omega_{\z_\body}$, the forward-arc motion primitive is computed as a polynomial
function of time generated using the following high-order constraints:
\begin{equation}
  \begin{aligned}
    \dot{\state}_{\tau} &= [v_{\x_\body}\cos{\theta}, v_{\x_\body} \sin{\theta}, v_{\z_\body}, \omega_{\z_\body}]\\
    {\state}^{(n)}_{\tau} &= \mathbf{0} \text{ for }n = 2, 3, 4
  \end{aligned}
\end{equation}
where $\{.\}^{(n)}$ denotes the $n^{\text{th}}$ time derivative, and $\tau$ is
the specified duration of the motion primitive. Motion primitives in the
$\y_\body$ direction can also be obtained by replacing $v_{\x_\body}$ by
$v_{\y_\body}$ in the above constraints. Later, we will use a combination of
these directions to define the action space for the exploration planner that
can operate with either a LiDAR or depth camera.

\subsubsection{Motion Primitive Library (MPL)}
A motion primitive library (MPL) is a collection of forward-arc motion
primitives generated using a user-specified discretization of the robot's
linear and angular velocities~\citep{yang2017framework}. Let $\action =
\{v_{\x_\body}, v_{\y_\body}, v_{\z_\body}, \omega_{\z_\body}\}$ be an action
set that is generated with user-specified maximum velocity bounds in the
$\x_\body - \y_\body$ plane and the $\z_\body$ direction. The motion primitive
library is then given by the set (\cref{subfig:single_mpl}):
\begingroup\makeatletter\def\f@size{8.5}\check@mathfonts
\def\maketag@@@#1{\hbox{\m@th\normalsize\normalfont#1}}
\begin{align}
  \moprimlib = \{ \moprimpseudo(\action_{jk}, \tau) \mid
  \|[v_\x, v_\y]\| \leq  \maxvelocity, \| v_\z \| \leq \verticalvelocity,
  \| \omega \| \leq \maxyawrate \}\label{eq:mpl}
\end{align}
\endgroup
where $j \in [1, N_{\omega}]$ and $k \in [1, N_{\z}]$ define the
action discretization for one particular primitive.

For each MPL, an additional MPL containing stopping trajectories
at any state $\state_t$ can be generated by fixing the desired
end point velocity to zero, $\dot{\state}_\tau = 0$. These stopping
trajectories are scheduled one planning round away from the starting
time of the planning round. These trajectories help ensure safety
in case the planner fails to compute an optimal action.

\subsubsection{Designing the Action Space} The final action space,
$\actionspace$, is a collection of MPLs selected according to three criteria:
(1) information gain rate, (2) safety, and (3) compute limitations.
Prior work~\citep{Michael-RSS-19} provides such a design for a
$360\si{\degree}$ FoV sensor (LiDAR).  \citet{goel2019fsr} present an analysis
on how these three factors influence $\actionspace$ for a limited FoV
depth sensor. This work extends~\citep{Michael-RSS-19} using the analysis
in~\citep{goel2019fsr}, yielding a motion planner amenable for exploration with
either a LiDAR or a depth sensor and that ensures similar exploration
performance in either case (see~\cref{sec:results}).

\paragraph{Action Space for $360\si{\degree}$ FoV Sensors} $360\si{\degree}$
FoV sensors are advantageous in an exploration scenario for three reasons:
(1) $360\si{\degree}$ depth data from the sensor allows for visibility
in all azimuthal directions, (2) a larger volume is explored per unit range when
compared to a limited FoV sensor, and (3) yaw in-place motion does not help
gain information. (1) enables backward and sideways motion in
the action space $\actionspace$ without sacrificing safety
(\cref{subfig:prims_lidar}). (2) influences the entropy
reduction: for the same trajectory, a sensor with a larger FoV will explore
more voxels compared to the limited FoV case. (3) reduces the
number of motion primitive libraries in the action space to yield increased
planning frequency. An example of an action space designed while considering
these factors is presented in~\citep{Michael-RSS-19} and the corresponding
parameters are shown in~\cref{stab:lidar}.

These factors indicate that the same action space $\actionspace$ cannot be used
for limited FoV cameras if comparable exploration performance is to be maintained.
This motivates the need for an alternate and informed action design for the
limited FoV cameras.

\begin{table}[t]
  \centering
  \adjustbox{valign=b}{%
  \subfloat[\label{stab:lidar}\textbf{LiDAR}]%
  {\begin{tabular}{|L|MLL|}
    \hline
    \textbf{MPL ID} & \textbf{Vel., Time} & $\numomega$, $\numz$ &
  $\numprimitives$\\
    \hline
    \num{1} & $v_{\x_\body}, \tau$ & \num{3}, \num{5} & \num{15}\\
    \num{2} & $v_{\x_\body}, 2\tau$ & \num{3}, \num{5} & \num{15}\\
    \num{3} & $-v_{\x_\body}, \tau$ & \num{3}, \num{5} & \num{15}\\
    \num{4} & $-v_{\x_\body}, 2\tau$ & \num{3}, \num{5} & \num{15}\\
    \hline
  \end{tabular}
  }}%
  \adjustbox{valign=b}{%
  \subfloat[\label{stab:tof}\textbf{Depth Camera}]%
  {\begin{tabular}{|L|MLL|}
    \hline
    \textbf{MPL ID} & \textbf{Vel., Time} & $\numomega$, $\numz$ &
  $\numprimitives$\\
    \hline
    \num{1} & $v_{\x_\body}, \tau$ & \num{3}, \num{5} & \num{15}\\
    \num{2} & $v_{\x_\body}, 2\tau$ & \num{3}, \num{5} & \num{15}\\
    \num{3} & $v_{\y_\body}, \tau$ & \num{3}, \num{5} & \num{15}\\
    \num{4} & $v_{\y_\body}, 2\tau$ & \num{3}, \num{5} & \num{15}\\
    \num{5} & $-v_{\y_\body}, \tau$ & \num{3}, \num{5} & \num{15}\\
    \num{6} & $-v_{\y_\body}, 2\tau$ & \num{3}, \num{5} & \num{15}\\
    \num{7} & $\omega_{\z_\body}, \tau$ & \num{1}, \num{5} & \num{5}\\
    \hline
  \end{tabular}}}
  \caption{
  \label{tab:action_space}
    Discretization used to construct the action space $\actionspace$ for the simulation experiments
    for \protect\subref{stab:lidar} LiDAR and \protect\subref{stab:tof} depth camera cases. Total number
    of primitives for a MPL are denoted by $\numprimitives = \numomega \cdot \numz$.
    The base duration $\tau$ was kept at $\SI{3}{\second}$ for all experiments.}
  \vspace{-0.5cm}
\end{table}

\paragraph{Action Space for Limited FoV Sensors}
\citet{goel2019fsr} show that for an exploration planner using a limited FoV
sensor, the design of the action space $\actionspace$ can be informed by the
sensor model. The authors consider a depth sensor to design
$\actionspace$ by incorporating the sensor range and FoV, among other factors.
This work follows a similar approach yielding an action space that contains MPLs
in both the $\x_\body$ and $\y_\body$ directions (\cref{subfig:prims_tof}). The
parameters to construct the MPL collection comprising $\actionspace$ are shown
in \cref{stab:tof}. Note that there is an additional MPL corresponding to a
yaw-in-place motion, unlike the $360\si{\degree}$ FoV case, to compensate for
the limited FoV of the depth camera. For further detail on how to
obtain these parameters, please refer to~\citep{goel2019fsr, Goel2019}.

\subsection{Information-Theoretic Objective}\label{sssec:information_theory}
The action selection policy uses CSQMI as the information-theoretic objective
to maximize the information gain over time. CSQMI is computed at $k$ points
along the primitive $\moprimpseudo$, and the sum is used as a metric to measure
the expected local information gain for a candidate action
$\informationreward$. However, this design may result in myopic
decision-making.
\clearpage
\begin{figure*}
  \centering
  \adjustbox{valign=b}{%
  \begin{minipage}{0.75\linewidth}
  \addtocounter{subfigure}{-1}
  \subfloat{%
    \ifthenelse{\equal{\arxivmode}{true}}%
    {\includegraphics[trim=0.1cm 0.1cm 0.1cm 0cm,clip]{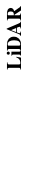}}%
    {\trimbox{0.35cm 0.1cm 0cm 0cm}{\input{results/sim/lidar3d_header.tex}}}%
  }%
  \subfloat[\label{sfig:6a}]{%
    \ifthenelse{\equal{\arxivmode}{true}}%
    {\includegraphics[trim=0.1cm 0.1cm 0cm 0cm,clip]{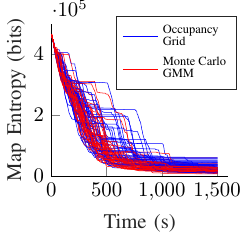}}%
    {\trimbox{0.35cm 0.1cm 0cm 0cm}{\input{results/sim/lidar3d_entropy_all_trials.tex}}}%
  }%
  \subfloat[\label{sfig:6b}]{%
    \ifthenelse{\equal{\arxivmode}{true}}%
    {\includegraphics[trim=0.1cm 0.1cm 0cm 0cm,clip]{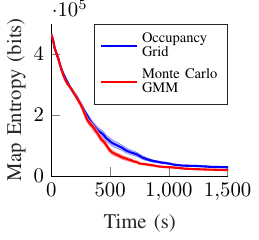}}%
    {\trimbox{0.2cm 0.1cm 0cm 0cm}{\input{results/sim/lidar3d_entropy.tex}}}%
  }%
  \subfloat[\label{sfig:6c}]{%
    \ifthenelse{\equal{\arxivmode}{true}}%
    {\includegraphics[trim=0.1cm 0.1cm 0cm 0cm,clip]{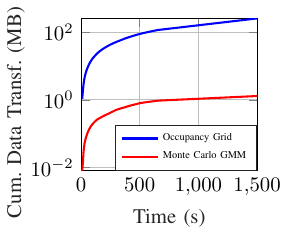}}%
    {\trimbox{0.2cm 0.05cm 0cm 0.1cm}{\input{results/sim/figure9c.tex}}}%
  }\\

  \addtocounter{subfigure}{-1}
  \subfloat{%
    \ifthenelse{\equal{\arxivmode}{true}}%
    {\includegraphics[trim=0.1cm 0.1cm 0.175cm 0cm,clip]{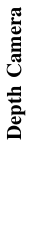}}%
    {\trimbox{0.35cm 0.1cm 0cm 0cm}{\input{results/sim/tof_header.tex}}}%
  }%
  \subfloat[\label{sfig:6d}]{%
    \ifthenelse{\equal{\arxivmode}{true}}%
    {\includegraphics[trim=0.1cm 0.1cm 0cm 0cm,clip]{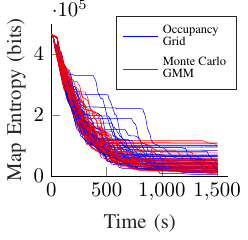}}%
    {\trimbox{0.35cm 0.05cm 0cm 0cm}{\input{results/sim/tof_entropy_all_trials.tex}}}%
  }%
  \subfloat[\label{sfig:6e}]{%
    \ifthenelse{\equal{\arxivmode}{true}}%
    {\includegraphics[trim=0.1cm 0.1cm 0cm 0cm,clip]{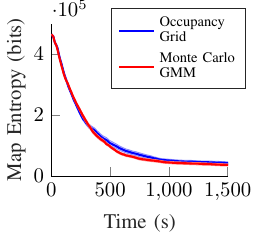}}%
    {\trimbox{0.2cm 0.05cm 0cm 0cm}{\input{results/sim/tof_entropy.tex}}}%
  }%
  \subfloat[\label{sfig:6f}]{%
    \ifthenelse{\equal{\arxivmode}{true}}%
    {\includegraphics[trim=0.1cm 0.1cm 0cm 0cm,clip]{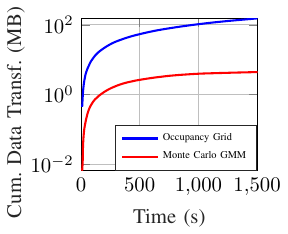}}%
    {\trimbox{0.2cm 0.05cm 0cm 0cm}{\input{results/sim/figure9f.tex}}}%
  }\\
  \end{minipage}%
    \begin{minipage}{0.22\linewidth}
      \subfloat[\label{sfig:6g}]{\vizbox{\vizfbox}{\includegraphics[width=\linewidth,trim=0 20 0 10,clip]{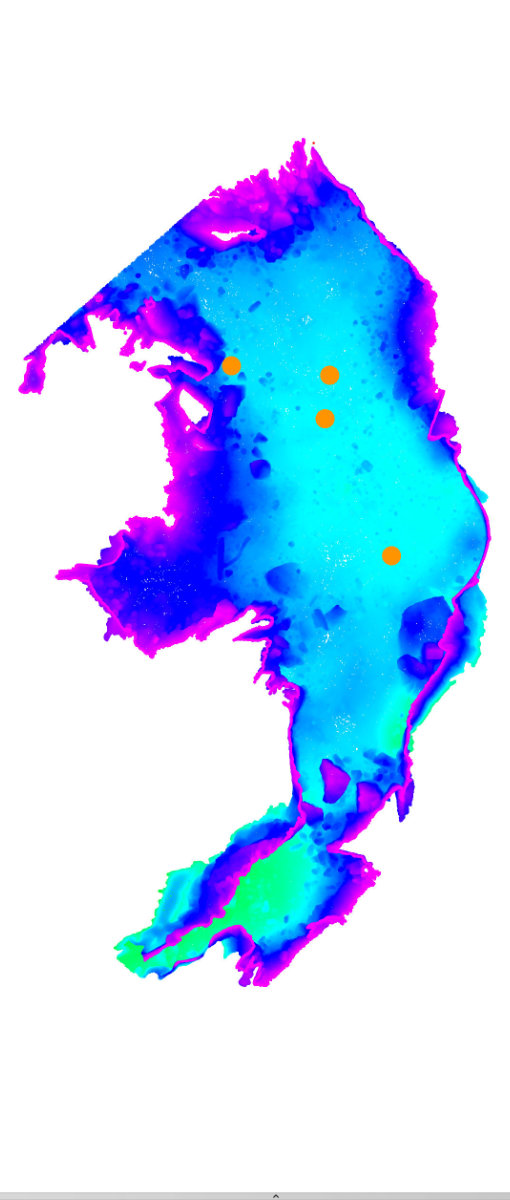}}}
    \end{minipage}}
  \caption{Exploration statistics for simulation experiments. The
    first row of results pertains to the LiDAR sensor model and the second
    row to the depth camera sensor model. \label{fig:sim_results}
    \protect\subref{sfig:6a} and \protect\subref{sfig:6d} illustrate the
    map entropy over time for 160 trials (80 trials per sensor model and
    40 trials per mapping method),~\protect\subref{sfig:6b}
    and~\protect\subref{sfig:6e} illustrate the average map entropy over time
    for each method. Although both methods achieve similar entropy
    reduction, MCG uses significantly less memory according to the average
    cumulative data transferred shown in~\protect\subref{sfig:6c}
    and~\protect\subref{sfig:6f}. When the LiDAR sensor model is used, the
    average cumulative data transferred at the end of \SI{1500}{\second} is
    \SI{1.3}{\mega\byte} for the MCG approach and
    \SI{256}{\mega\byte} for the OG approach. When the depth camera sensor
    model is used, the average cumulative data transferred at the end of \SI{1500}{\second} is
    \SI{4.4}{\mega\byte} for the MCG approach and \SI{153}{\mega\byte} for
    the OG approach. The MCG
    method represents a decrease of approximately one to two orders of
    magnitude as compared to the OG method for the LiDAR
    and depth camera sensor models, respectively. The experiments are
    conducted in the simulated cave environment shown in~\cref{sfig:6g}. The
    four starting positions are shown as orange dots.}

   \centering
  \subfloat[\label{sfig:10a}Simulated Cave Environment]{\includegraphics[width=0.22\linewidth,trim=90 50 90 50,clip]{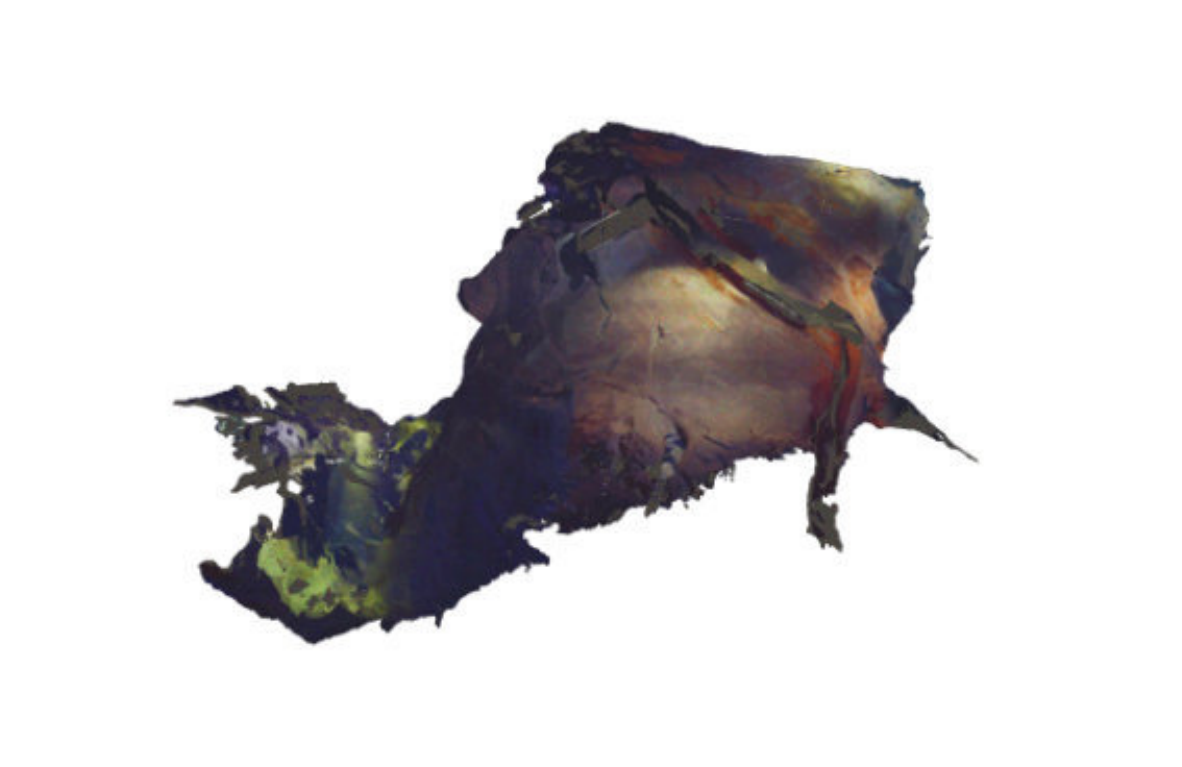}}
  \addtocounter{subfigure}{-1}
  \subfloat{%
    \ifthenelse{\equal{\arxivmode}{true}}%
    {\includegraphics[trim=0.1cm 1.25cm 0cm 0cm,clip]{./tikzext/lidar3d_header.pdf}}%
    {\trimbox{0.4cm 1.25cm 0cm 0cm}{\input{results/sim/lidar3d_header.tex}}}%
  }%
  \subfloat[\label{sfig:gmm_map_lidar}GMM map (1$\sigma$ covariances)]{\includegraphics[width=0.23\linewidth,trim=0 0 0 0,clip]{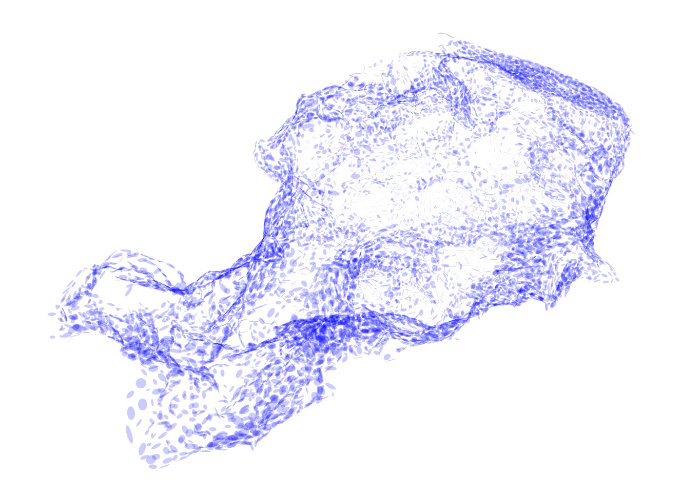}}
  \subfloat[\label{sfig:resampled_lidar}Resampled pointcloud]{\includegraphics[width=0.23\linewidth,trim=0 0 0 0,clip]{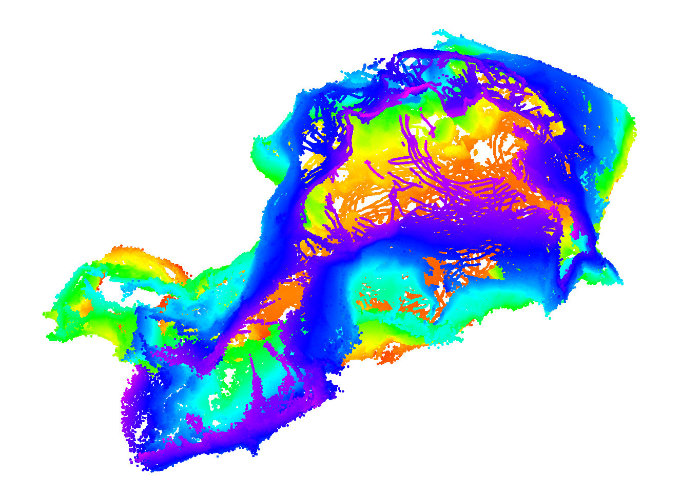}}
  \subfloat[\label{sfig:dense_voxel_lidar}Dense voxel map]{\includegraphics[width=0.23\linewidth,trim=0 0 0 0,clip]{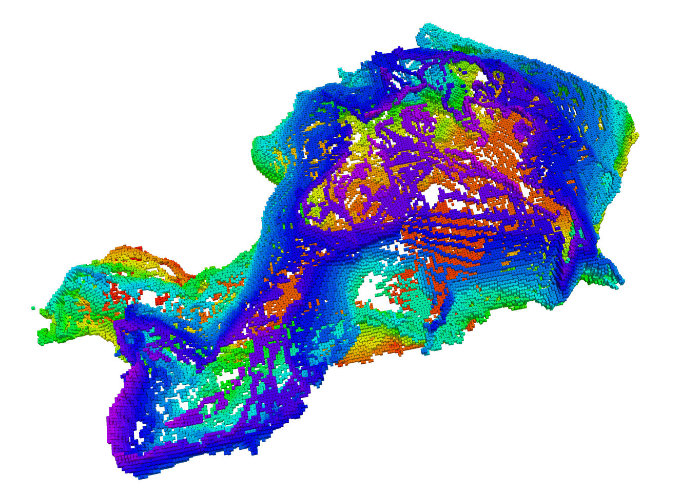}}\\

  \subfloat[\label{sfig:10e}Points from the PLY]{\includegraphics[width=0.2\linewidth,trim=0 0 0 0,clip]{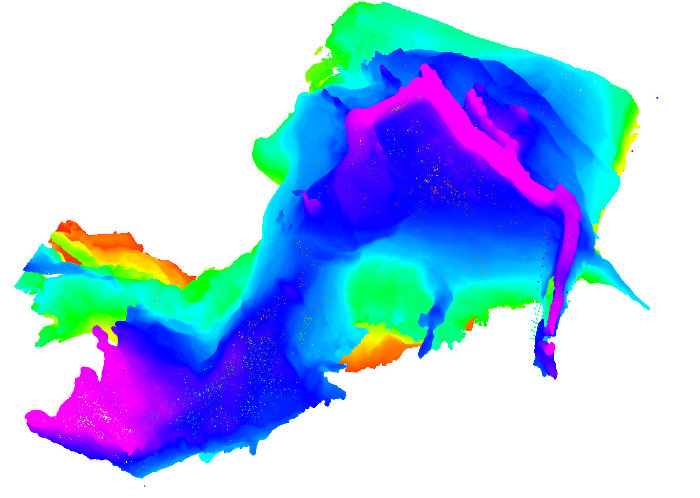}}%
  \addtocounter{subfigure}{-1}
  \subfloat{%
    \ifthenelse{\equal{\arxivmode}{true}}%
    {\includegraphics[trim=-0.3cm 1.25cm 0cm 0cm,clip]{./tikzext/tof_header.pdf}}%
    {\trimbox{0.0cm 1.25cm 0cm 0cm}{\input{results/sim/tof_header.tex}}}%
  }%
  \subfloat[\label{sfig:gmm_map_tof}GMM map (1$\sigma$ covariances)]{\includegraphics[width=0.23\linewidth,trim=0 0 0 0,clip]{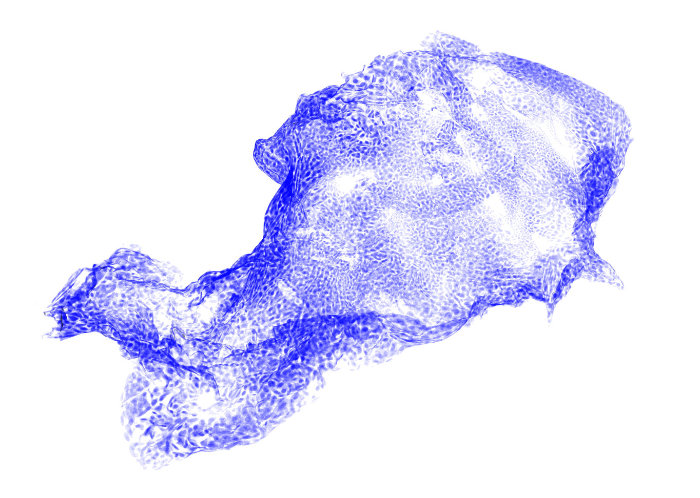}}
  \subfloat[\label{sfig:resampled_tof}Resampled pointcloud]{\includegraphics[width=0.23\linewidth,trim=0 0 0 0,clip]{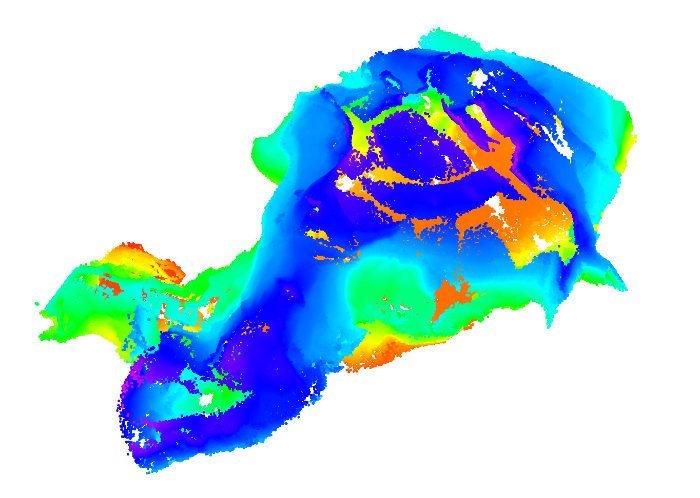}}
  \subfloat[\label{sfig:dense_voxel_tof}Dense voxel map]{\includegraphics[width=0.23\linewidth,trim=0 0 0 0,clip]{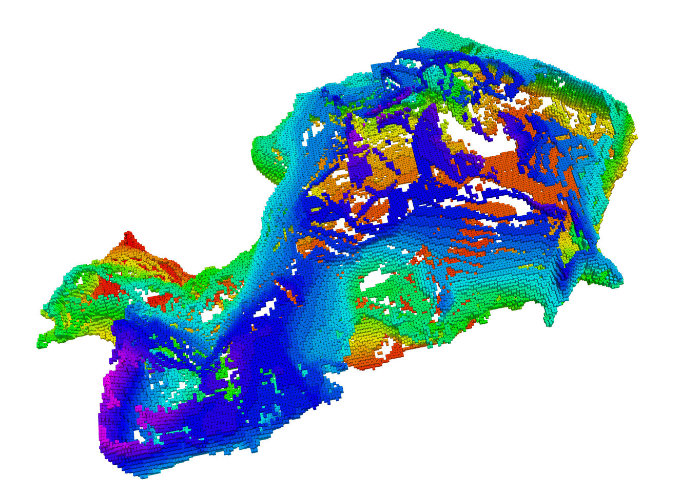}}\\

  \caption{\label{fig:10} The colorized mesh used in
    simulation experiments is shown in~\protect\subref{sfig:10a}
    and produced from FARO scans of a cave in West Virginia.
    After \SI{1500}{\second} of exploration with a LiDAR sensor model,
    the resulting~\protect\subref{sfig:gmm_map_lidar} MCG map is shown with 1$\sigma$ covariances
    and densely resampled with \SI{1e6}{} points to obtain the reconstruction shown
    in~\protect\subref{sfig:resampled_lidar}.
    \protect\subref{sfig:dense_voxel_lidar} illustrates the dense voxel map produced after a
    \SI{1500}{\second} trial with \SI{20}{\centi\meter} voxels.
    \protect\subref{sfig:10e} illustrates the pointcloud from the mesh
    shown in~\protect\subref{sfig:10a}.
    \protect\subref{sfig:gmm_map_tof} illustrates the MCG map
    with 1$\sigma$ covariances, which is densely resampled with
    \SI{1e6}{} points, to obtain the reconstruction shown in~\protect\subref{sfig:resampled_tof}.
    \protect\subref{sfig:dense_voxel_tof} illustrates the dense voxel map
    with \SI{20}{\centi\meter} voxels
    after \SI{1500}{\second} of exploration with the depth camera sensor model. The
    reconstruction accuracy for \protect\subref{sfig:resampled_lidar}, \protect\subref{sfig:dense_voxel_lidar},
    \protect\subref{sfig:resampled_tof}, and \protect\subref{sfig:dense_voxel_tof}
    are shown in~\cref{tab:reconstruction_accuracy}.
    All pointclouds shown are colored from red to purple according to z-height.}
\end{figure*}
\newpage


Therefore, frontiers are also incorporated to model the global spatial
distribution of information~\citep{yamauchi1997}. This global reward, denoted by
$\viewdistancereward$, is calculated based on the change in distance towards a
frontier along a candidate action. Using the node state $\state_{0}$, end point
state $\state_{\tau}$, and a distance field constructed based on the position of
the frontiers, this reward can be calculated as $\viewdistancereward =
d(\state_{0}) - d(\state_{\tau})$, where $d(\state_t)$ denotes the distance to
the nearest voxel in the distance field from state
$\state_t$~\citep{corah2018ral}.

\begin{algorithm}[t]
  \caption{Overview of Action Selection for Exploration}
  \label{alg:overall_action_selection}
  \begin{minipage}{\linewidth}
    \small
    \begin{algorithmic}[1]
      \State \textbf{input}: $\actionspace$, $\safespace$
      \State \textbf{output}: $\bestaction$ \Comment best action
      \For{$\moprimlib \in \actionspace$}
        \For{$\moprimpseudo \in \moprimlib$}
        \State \emph{feasible} $\gets$ \Call{SafetyCheck}{$\moprimpseudo$, $\stopmoprim$, $\safespace$}\label{line:safety_check}
        \If{\emph{feasible}}
        \State $\informationreward \gets$ \Call{InformationReward}{$\moprimpseudo$}\label{line:information_reward}
          \State $\viewdistancereward \gets$ \Call{FrontierDistanceReward}{$\moprimpseudo$}\label{line:distance_reward}
        \Else
          \State $\informationreward \gets 0.0$, $\viewdistancereward \gets 0.0$
        \EndIf
        \EndFor
      \EndFor
      \State \Return $\bestaction \gets \argmax\limits_{\moprimpseudo \in \actionspace}
      [\informationreward + \viewdistancereward]$\label{line:return}
    \end{algorithmic}
  \end{minipage}
\end{algorithm}

\subsection{Action Selection}\label{sssec:action_selection}
Using the rewards described in the preceding section, the objective for the
motion planner is defined as follows~\citep{goel2019fsr, corah2018ral}:
\begin{align}
  \begin{split}
    \argmax\limits_{\moprimpseudo}~& \informationreward + \terminalweight \, \viewdistancereward \\
    \mbox{s.t.}~&
    \moprimpseudo\in\actionspace
  \end{split}
  \label{eq:objective}
\end{align}
where $\terminalweight$ is a weight that adjusts the contribution of the
frontier distance reward. Recall, the goal is to maximize this reward function
in real-time on a compute-constrained aerial platform.  Previous
information-theoretic approaches that construct a tree and use a finite-horizon
planner either do not use a global heuristic~\citep{tabib2016iros} or are not
known to be amenable for operation on compute-constrained
platforms~\citep{corah2018ral}. In this work, a single-step planner is used
with the action space $\actionspace$ consisting of motion primitives of varying
duration for real-time performance (see~\cref{tab:action_space}).  Due to this
choice, the planner computes rewards over candidate actions that
extend further into the explored map from the current position. In this manner,
longer duration candidate actions provide a longer lookahead than the case when
all candidate actions are of the same duration, even in single-step planning
formulations (see~\cref{tab:action_space}).

The action selection procedure is detailed
in~\cref{alg:overall_action_selection}. For every candidate action
$\moprimpseudo$ in the action space $\actionspace$, a safety check procedure is
performed to ensure that this candidate and the associated stopping action
($\stopmoprim$) are dynamically feasible and lie within free space $\safespace$
(\cref{line:safety_check}).  The free space check is performed using a
Euclidean distance field created from locations of occupied and unknown spaces
in the robot's local map given a fixed collision radius~\citep{corah2018auro}.
Checking that the stopping action is feasible ensures that the planner
never visits an inevitable collision state, which is essential for safe
operation~\citep{Janson2018}. If the action is feasible, the local
information reward ($\informationreward$,~\cref{line:information_reward}) and
frontier distance reward ($\viewdistancereward$,~\cref{line:distance_reward})
are determined as described in~\cref{sssec:information_theory}. The planner
then returns the action with the best overall reward (\cref{line:return}).
\begin{table}[b]
  \centering
  \begin{tabular}{c|cc|cc}
    \toprule
    \textbf{Approach} & \multicolumn{2}{c}{\textbf{LiDAR}}                          & \multicolumn{2}{c}{\textbf{Depth Camera}}            \\
    \midrule
            & \textbf{Mean (\SI{}{\meter})} & \textbf{Std (\SI{}{\meter})} & \textbf{Mean (\SI{}{\meter})} & \textbf{Std (\SI{}{\meter})}  \\
    \midrule
    MCG     & \SI{1.8e-02}{}                & \SI{2.5e-02}{}               & \SI{1.3e-02}{}                & \SI{1.9e-02}{}                \\
    OG      & \SI{6.2e-02}{}                & \SI{3.9e-02}{}                & \SI{6.3e-02}{}               & \SI{3.9e-02}{}               \\
    \bottomrule
  \end{tabular}
  \caption{\label{tab:reconstruction_accuracy} Reconstruction error for~\cref{fig:10}. The error is calculated as the PointCloud-to-Mesh distance between the environment reconstructions and mesh.}
\end{table}


\section{Experimental Design and Results\label{sec:results}}
This section details the experimental design to validate the
approach.  Results are reported
for both real-time simulation trials and field tests in caves.
The following shorthand is introduced for this section only: MCG
will refer to the Monte Carlo GMM mapping approach and OG mapping will refer to
the Occupancy Grid mapping approach. The mapping and planning software is run
on an embedded Gigabyte Brix 8550U with eight cores and \SI{32}{\giga\byte}
RAM, for both hardware and simulation experiments.  Simulation results are
presented for both LiDAR and depth camera sensor models, but hardware results
are reported only for the depth camera case\footnote{The prior work upon which
  this manuscript is developed leveraged a \SI{6.7}{\kilo\gram} aerial system with
  LiDAR.
To support improved experimental convenience, an alternative platform was
developed that results in lower size, weight, and power consumption as
compared to the previous platform. For the LiDAR system hardware results,
please see \citep{Michael-RSS-19}.}. Unless otherwise noted, the parameters for
simulation and hardware experiments are equal.

\subsection{Comparison Metrics}
To calculate the memory requirements for the OG mapping approach, the incremental
OG map is transmitted as a changeset pointcloud where each point consists of 4
floating point numbers: $\{x,y,z,\text{logodds}\}$. The changeset is computed
after insertion of every pointcloud. A floating point number is assumed to be
four bytes, or 32 bits. For the MCG approach, the cumulative data transferred
is computed by summing the cost of transmitted GMMs. Each mixture component
is transmitted as 10 floating
point numbers: six numbers for the symmetric covariance matrix,
three numbers for the mean, and one number for the mixture
component weight. One additional number is stored per GMM that represents
the number of points from which the GMM was learned. The transform
  between the sensor origin reference frame and the global reference frame is stored
  for each GMM using six numbers to represent the three translational and three
  rotational degrees of freedom. To ensure a fair comparison of exploration
performance between the two approaches, a global occupancy grid serves as a referee and is
maintained in the background with a voxel resolution of $\SI{0.2}{\meter}$.
Exploration progress in simulation and hardware experiments
is measured using the map entropy.
The map entropy quantifies how the map's uncertainty changes over time
using the Shannon entropy of the global occupancy grid cells the robot could potentially
observe~\cite{charrow2015icra,cover2012elements}.

\subsection{Simulation Experiments\label{ssec:simulation_results}}
The exploration strategy is evaluated with 160 real-time
simulation trials over approximately 67 hours in a $\SI{30}{\meter}
\times \SI{40}{\meter} \times \SI{6}{\meter}$ environment constructed
from colorized FARO
pointclouds of a cave in West Virginia
(see~\cref{sfig:10a}). In each simulation, the multirotor robot
begins exploration from one of four pre-determined starting positions
and explores for \SI{1500}{\second}. Ten exploration tests for
  each of the four sensor configurations are run from each of the four starting
  positions, leading to a total of 160 trials. The end time of \SI{1500}{\second}
is empirically set based on the total time required to fully explore the cave.  Note that ground
truth state estimates are used for these simulation experiments, while
the hardware experiments in~\cref{ssec:hardware_results} rely on
visual-inertial odometry (see~\cref{ssec:vins}).
The reconstruction error
(\cref{tab:reconstruction_accuracy}
and~\cref{sfig:resampled_lidar,sfig:dense_voxel_lidar,sfig:resampled_tof,sfig:dense_voxel_tof})
is computed as pointcloud-to-mesh distances between reconstructed pointclouds from
each trial and the environment mesh. In the case the MCG approach, the GMM map is
densely resampled to produce a pointcloud. For the OG case, the
occupancy grid map is converted to a pointcloud by assuming the points
to be at the center of each voxel.

\subsubsection{LiDAR Simulations}
The LiDAR has a max range of $\SI{5.0}{\meter}$ and operates at $\SI{10}{\hertz}$ for
all simulation experiments. The motion planning parameters used in the action space design
are shown in \cref{stab:lidar}. For all simulation trials, the maximum speed
in the $\xyplane$ plane is $\| \maxvelocity \| = \SI[per-mode=symbol]{0.75}{\meter\per\second}$,
the maximum speed along the $\zaxis$ axis is $\verticalvelocity = \SI[per-mode=symbol]{0.5}{\meter\per\second}$,
and the maximum yaw rate is $\maxyawrate =\SI[per-mode=symbol]{0.25}{\radian\per\second}$. CSQMI
is computed at the end point of the candidate action ($k = 1$). $\lambda = 5$ and $n_f = 2$ for
all simulations and hardware trials.

The simulation trials demonstrate that MCG achieves similar exploration performance
as OG, which indicates that the approximations made
by the former enables real-time performance without compromising exploration or
map reconstruction quality (\cref{sfig:6b,sfig:resampled_lidar}). \Cref{sfig:6c} depicts the
cumulative data that must be
transferred to reproduce the OG and MCG maps remotely. After \SI{1500}{\second},
transferring the MCG map requires \SI{1.3}{\mega\byte} as compared to
\SI{256}{\mega\byte} to incrementally transfer the OG map. The MCG approach
significantly outperforms the OG approach in terms of cumulative data
transfer requirements.
A representative example of the reconstructed GMM map for one trial
from~\cref{sfig:6a} is shown in~\cref{sfig:gmm_map_lidar}.
Resampling \SI{1e6}{} points from the distribution yields the map shown
in~\cref{sfig:resampled_lidar}. The MCG approach has
lower average reconstruction error as compared to the OG approach (\cref{sfig:dense_voxel_lidar})
as shown in~\cref{tab:reconstruction_accuracy}.

\subsubsection{Depth Camera Simulations}
The depth camera sensor model also has a max range of $\SI{5.0}{\meter}$ and operates at $\SI{10}{\hertz}$
for all simulation experiments. A collection
of motion primitive libraries used for the simulation experiments is shown in \cref{stab:tof}.
The velocity bounds for the simulation experiments are the same as in the LiDAR case.

\begin{figure*}
  \centering
  \subfloat[\label{sfig:fa_all_entropy}]{%
    \ifthenelse{\equal{\arxivmode}{true}}%
    {\includegraphics[]{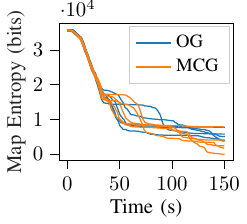}}%
    {\input{./flight_arena/figures/entropy_all_trials.tex}}%
  }%
  \subfloat[\label{sfig:fa_mean_entropy}]{%
    \ifthenelse{\equal{\arxivmode}{true}}%
    {\includegraphics[]{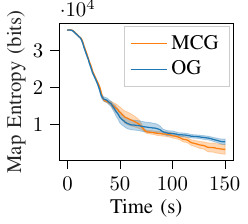}}%
    {\input{./flight_arena/figures/entropy_means.tex}}%
  }%
  \subfloat[\label{sfig:fa_all_comms}]{%
    \ifthenelse{\equal{\arxivmode}{true}}%
    {\includegraphics[]{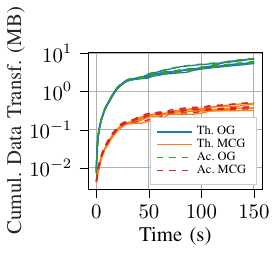}}%
    {\input{./flight_arena/figures/comms_all_trials.tex}}%
  }%
  \subfloat[\label{sfig:fa_mean_comms}]{%
    \ifthenelse{\equal{\arxivmode}{true}}%
    {\includegraphics[]{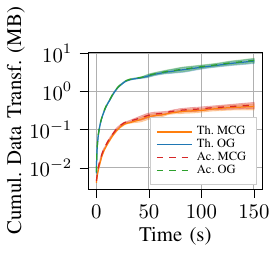}}%
    {\input{./flight_arena/figures/comms_means.tex}}%
  }\\
  \subfloat[\label{sfig:fa_flying_robot}]{%
    \ifthenelse{\equal{\arxivmode}{true}}%
    {\includegraphics[height=4cm,trim=300 0 20 50,clip]{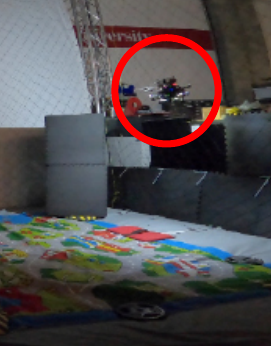}}%
    {\includegraphics[height=4cm,trim=300 0 20 50,clip]{./flight_arena/affinity/flight6-gmm-low-res.eps}}%
  }%
  \subfloat[\label{sfig:fa_reconstructed_map}]{%
    \ifthenelse{\equal{\arxivmode}{true}}%
    {\includegraphics[height=4cm,trim=100 0 20 50,clip]{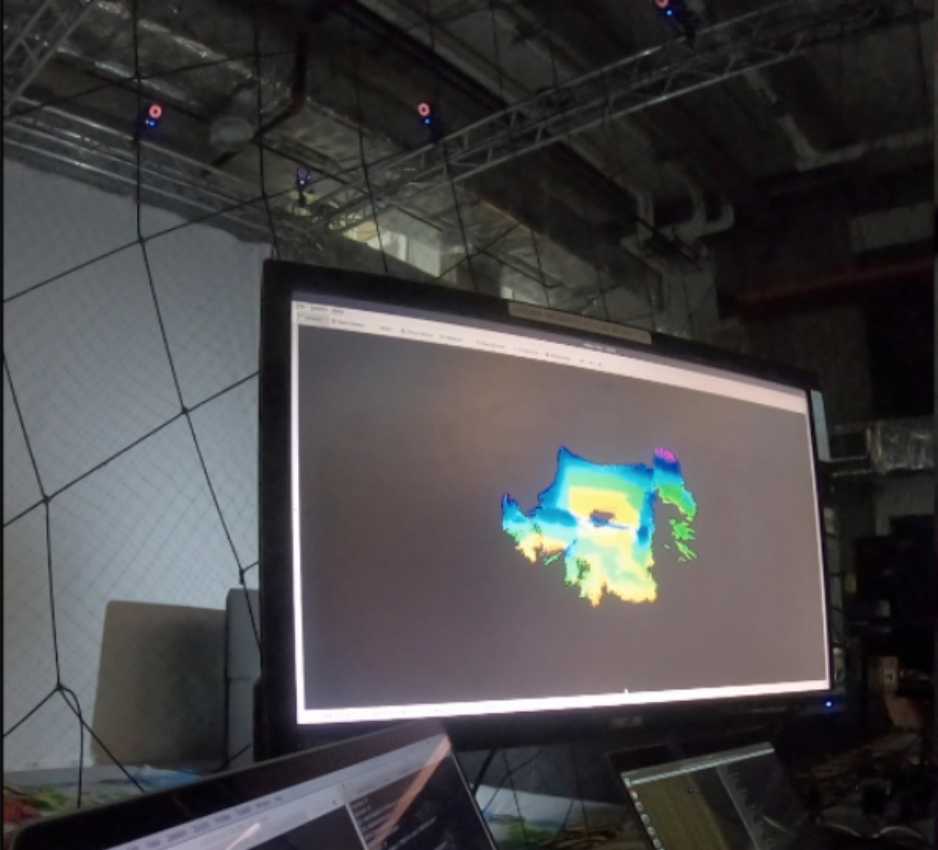}}%
    {\includegraphics[height=4cm,trim=100 0 20 50,clip]{./flight_arena/affinity/flight6-basestation-low-res.eps}}%
  }%
  \subfloat[\label{sfig:planning_times}]{%
    \ifthenelse{\equal{\arxivmode}{true}}%
    {\includegraphics[]{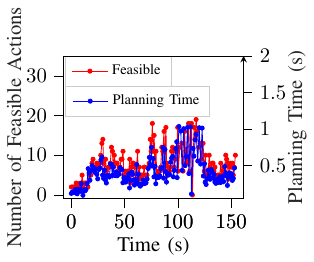}}%
    {\input{./flight_arena/figures/planning_time.tex}}%
  }%
  \subfloat[\label{sfig:octomap_comparison}]{%
    \ifthenelse{\equal{\arxivmode}{true}}%
    {\includegraphics[]{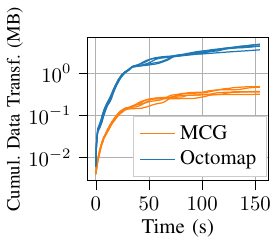}}%
    {\input{./flight_arena/figures/figure11h.tex}}%
  }\\
    \subfloat[\label{tab:fa_planning}Planning and mapping statistics for the MCG trials in the flight arena]{\input{./flight_arena/figures/planning_table.tex}}\\
    \subfloat[\label{tab:fa_mem_compute}Memory and compute statistics for the MCG trials in the flight arena]{\input{./flight_arena/figures/compute_table.tex}}
    \caption{\label{fig:flight_arena}Results of repeatability trials for the MCG and OG
approaches in a flight arena. \protect\subref{sfig:fa_all_entropy}
illustrates the entropy reduction for five trials for the MCG and OG methods.
\protect\subref{sfig:fa_mean_entropy} plots the standard error on top
of the mean line. The cumulative data transferred is provided for each
approach in~\protect\subref{sfig:fa_all_comms} with the mean and
standard deviation for the trials shown
in~\protect\subref{sfig:fa_mean_comms}. The theoretical (Th. OG and Th. MCG)
communications is compared to actual (Ac. OG and Ac. MCG) communications transmitted to the
base station using UDP.~\protect\subref{sfig:fa_flying_robot}
is a still image of the robot flying during one of the MCG trials (full video of the
trial may be found at \url{https://youtu.be/egwjv7YwHPE}) and~\protect\subref{sfig:fa_reconstructed_map}
illustrates the live map transmitted to the base station from the same
trial.~\protect\subref{sfig:planning_times} provides a plot of the number of
feasible actions in red with the planning time shown in blue.
~\protect\subref{sfig:octomap_comparison} Uses data from the MCG
flights and generates an Octomap in postprocessing to compare the
communications required. The Octomap performance is similar to that of
the OG approach. More details about this analysis is provided
in~\cref{ssec:flight_arena}.~\protect\subref{tab:fa_planning}
and~\protect\subref{tab:fa_mem_compute} provide timing, compute, and
memory statistics for each subsystem for each of the five MCG flights. The
figures reported in~\cref{tab:fa_mem_compute} are averages.}
\end{figure*}

\begin{figure*}[h]
  \centering
  \subfloat[\label{sfig:ghc_data_sent}]{%
    \ifthenelse{\equal{\arxivmode}{true}}%
    {\includegraphics[]{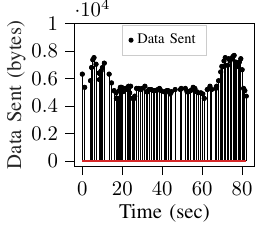}}%
    {\input{./flight_arena/figures/sent_comms.tex}}%
  }%
  \subfloat[\label{sfig:ghc_data_rec}]{%
    \ifthenelse{\equal{\arxivmode}{true}}%
    {\includegraphics[]{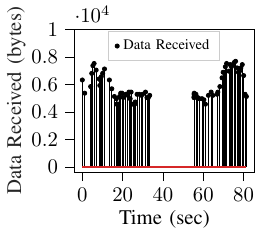}}%
    {\input{./flight_arena/figures/received_comms.tex}}%
  }%
  \subfloat[\label{sfig:ghc_resulting_map}]{%
    \ifthenelse{\equal{\arxivmode}{true}}%
    {\includegraphics[height=3.7cm,trim=50 0 100 0,clip]{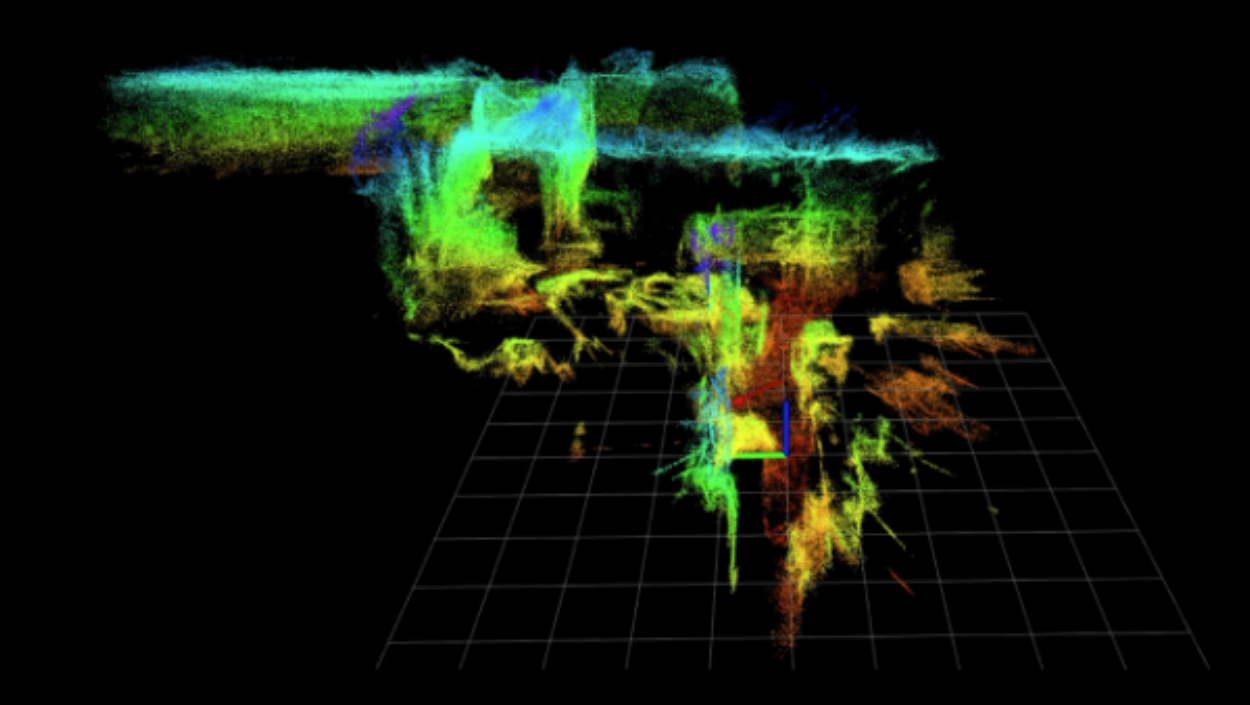}}%
    {\includegraphics[height=3.7cm,trim=50 0 100 0,clip]{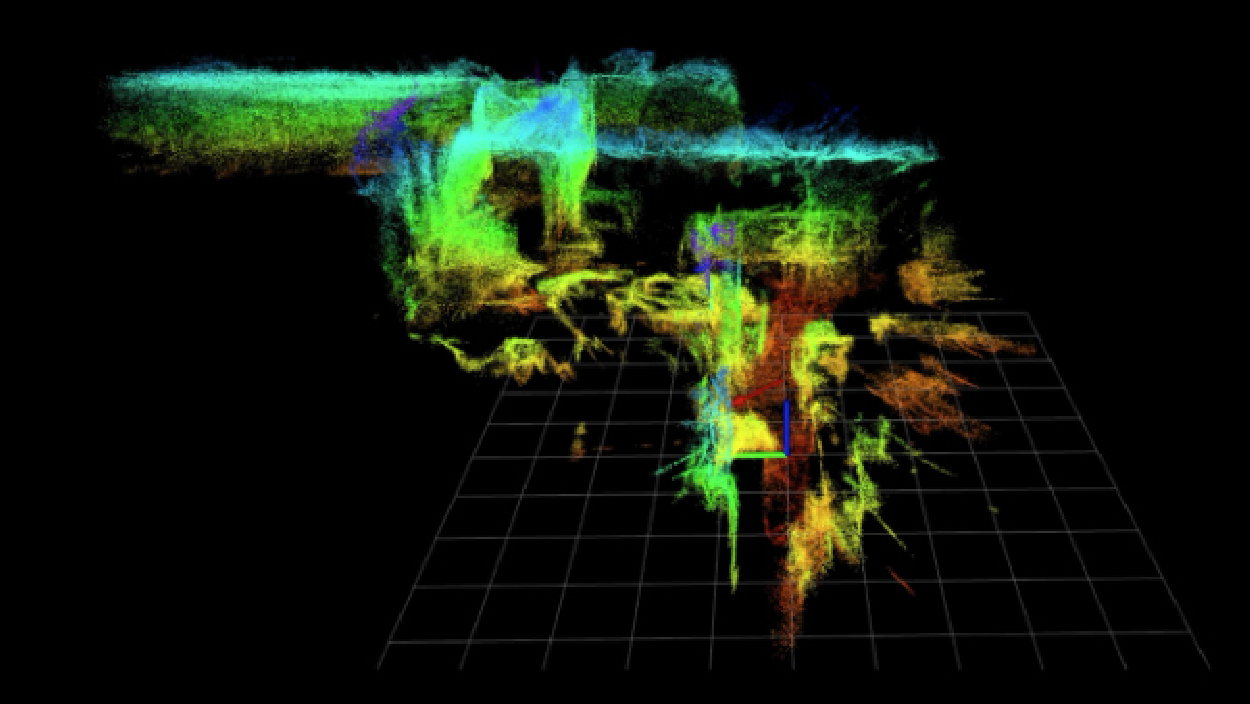}}%
  }%
  \subfloat[\label{sfig:ghc_helmet_cam}]{%
    \ifthenelse{\equal{\arxivmode}{true}}%
    {\includegraphics[height=3.7cm,trim=0 0 0 20,clip]{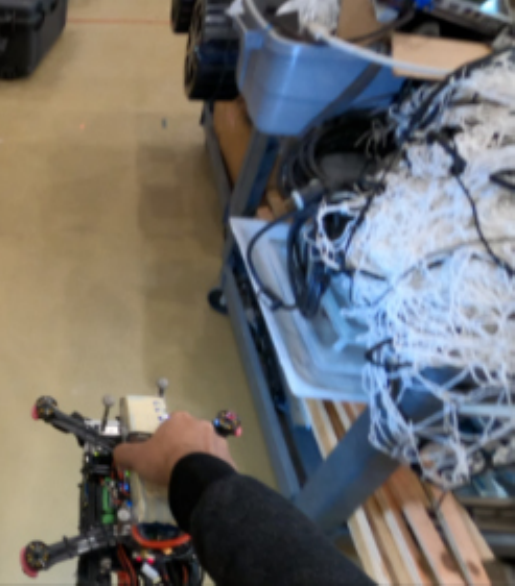}}%
    {\includegraphics[height=3.7cm,trim=0 0 0 20,clip]{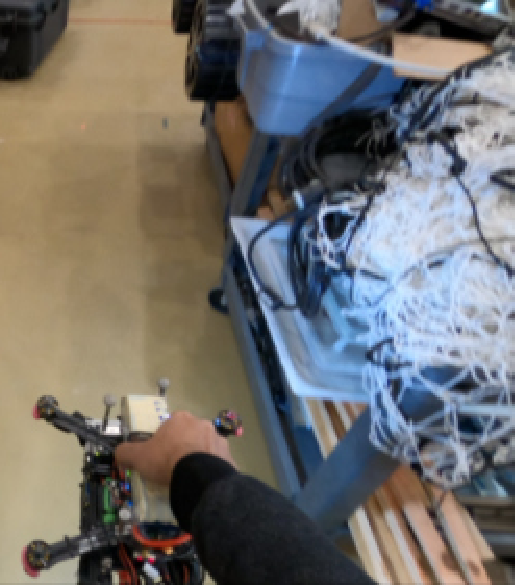}}%
  }%
\caption{\label{fig:dropouts}Results of forcing a communication dropout on the system. The
aerial system is carried through a research lab and down a hallway
away from the base station and router to force a communications
dropout. The accompanying video may be found at \url{https://youtu.be/UVn2BbMQRJg}.
~\protect\subref{sfig:ghc_data_sent} illustrates the data sent
from the robot and~\protect\subref{sfig:ghc_data_rec} is the data
received by the base station (note: the base station and robot do not have
their clocks synced).~\protect\subref{sfig:ghc_resulting_map} illustrates the
live map produced by the base station.~\protect\subref{sfig:ghc_helmet_cam}
illustrates a view of the aerial system at the start of the experiment from a
camera mounted on the operator's helmet.}
\end{figure*}


\begin{figure*}
  \centering
  \subfloat[\label{sfig:rocky0803_explores}]{\includegraphics[width=\linewidth,trim=0 0 0 0,clip]{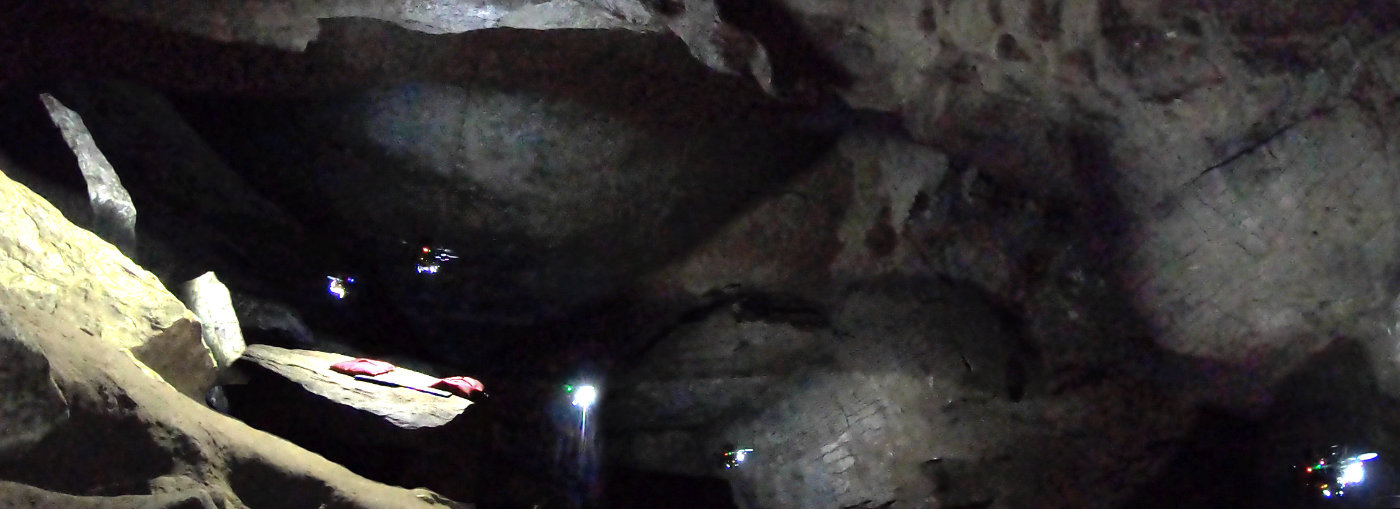}}\\
  \subfloat[\label{sfig:rocky0803}]{\vizbox{\vizfbox}{\includegraphics[width=0.33\linewidth,trim=35 0 5 0,clip]{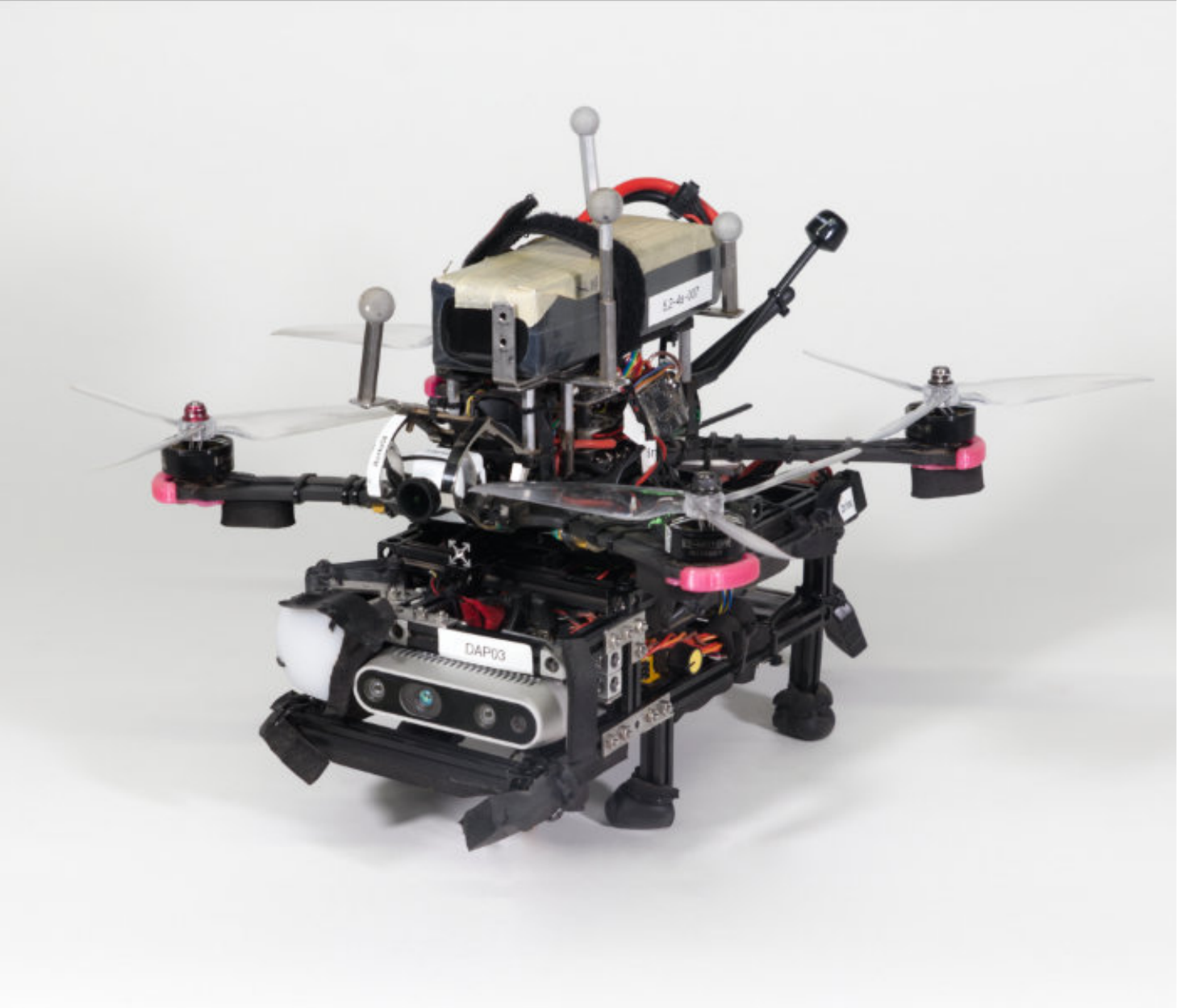}}}%
  \subfloat[\label{sfig:reconstruction_error}]{\vizbox{\vizfbox}{\includegraphics[width=0.33\linewidth,trim=0 0 0 0,clip]{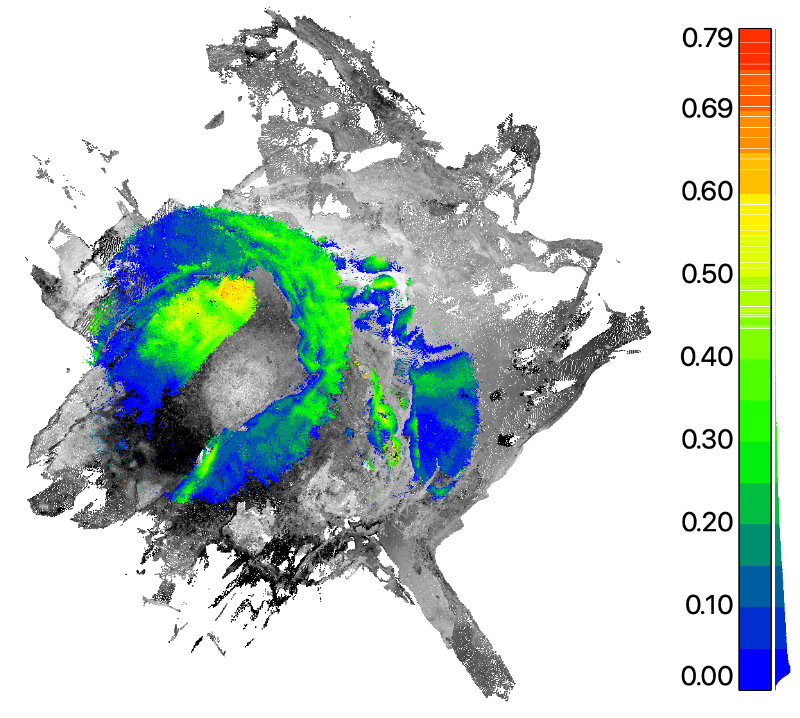}}}%
  \subfloat[\label{sfig:figure11d}]{\vizbox{\vizfbox}{\includegraphics[width=0.33\linewidth,trim=90 0 50 0,clip]{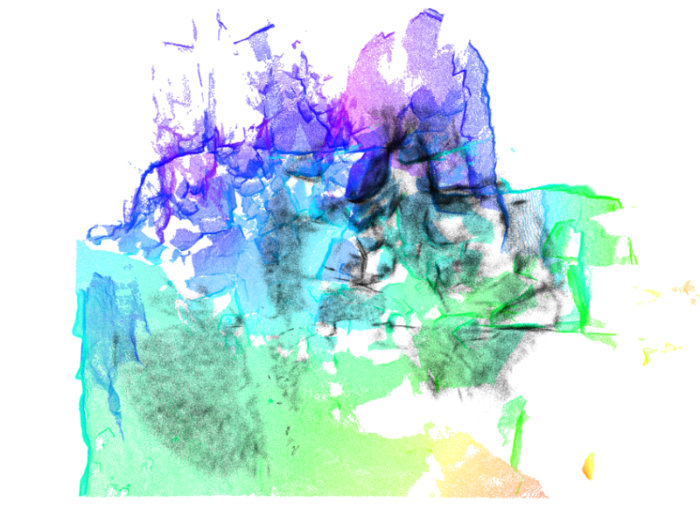}}}\\
  \subfloat[\label{sfig:laurel_caverns_entropy}]{%
    \ifthenelse{\equal{\arxivmode}{true}}%
    {\includegraphics[]{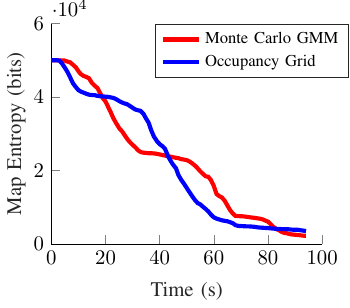}}%
    {\input{laurel_caverns/laurel_caverns_entropy.tex}}%
  }%
  \subfloat[\label{sfig:laurel_caverns_comms}]{%
    \ifthenelse{\equal{\arxivmode}{true}}%
    {\includegraphics[]{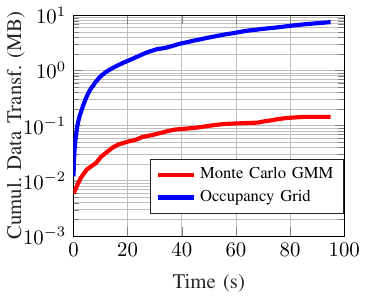}}%
    {\input{laurel_caverns/figure13f.tex}}%
  }%
  \subfloat[\label{sfig:laurel_caverns_rate}]{%
    \ifthenelse{\equal{\arxivmode}{true}}%
    {\includegraphics[]{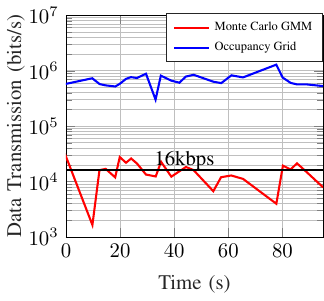}}%
    {\input{laurel_caverns/figure13g.tex}}%
  }\\
  \caption{\label{fig:laurel_caverns}
    \protect\subref{sfig:rocky0803_explores} A single aerial system
    explores the Dining Room of Laurel Caverns in Southwestern
    Pennsylvania. Still images of the robot exploring the environment
    are super-imposed to produce this figure.
    \protect\subref{sfig:rocky0803} The aerial
    system with dimensions $\SI{0.25}{\meter}\times\SI{0.41}{\meter}
    \times\SI{0.37}{\meter}$ including propellers carries a
    forward-facing Intel Realsense D435 for mapping and
    downward-facing global shutter MV Bluefox2 camera (not shown).
    The pearl reflective markers are used for testing in a
    motion capture arena but are not used during field
    operations to obtain hardware results.  Instead, a tightly-coupled
    visual-inertial odometry framework is used to estimate state
    during testing at Laurel Caverns.
    \protect\subref{sfig:reconstruction_error} illustrates the
    reconstruction error of the resampled GMM map as compared to
    the FARO map by calculating point-to-point distances.  The
    distribution of distances is shown on the right-hand side.
    The mean error is \SI{0.14}{\meter} with a standard deviation
    of \SI{0.11}{\meter}. In particular, there is misalignment in
    the roof due to pose estimation drift.
    \protect\subref{sfig:figure11d} A subset of the resampled GMM map (shown
    in black) is overlaid onto the FARO map (shown in colors ranging from red to purple)
    that displays the breakdown in the middle of the Dining Room.
    \protect\subref{sfig:laurel_caverns_entropy} The entropy
    reduction and
    \protect\subref{sfig:laurel_caverns_comms}
    cumulative data transferred for one trial for each of the
    Monte Carlo GMM mapping and OG mapping approaches are shown.
    The communication is a theoretical calculation -- not actual transmitted data. While the map entropy reduction
    for each approach is approximately similar, the GMM mapping
    approach transmits significantly less memory than the OG
    mapping approach (\SI{0.1}{\mega\byte} as compared to
    \SI{7.5}{\mega\byte}).~\protect\subref{sfig:laurel_caverns_rate}
    illustrates the bit rate for each approach in a semi-logarithmic plot where the vertical
    axis is logarithmic.  The black line illustrates how the approaches compare to 16kbps.
    For comparison, 16kbps
    is sufficient to transmit a low resolution (176 $\times$ 144 at 5 fps compressed to 3200 bit/frame)
    \emph{talking heads} video~\cite{389401,959054}.}
\end{figure*}
Similar to the LiDAR results, MCG outperforms OG in terms of memory
efficiency while maintaining similar exploration
performance. \Cref{sfig:6f} depicts the cumulative amount of data
transfer in this case. After \SI{1500}{\second}, transferring the
MCG map requires \SI{4.4}{\mega\byte} and
\SI{153}{\mega\byte} to incrementally transfer the OG
map. Because the LiDAR has a larger field of view and lower resolution
(i.e., fewer points for the same area of coverage) as compared to the
depth camera, the LiDAR sensor model covers more voxels than a depth camera
  sensor model with the same voxel resolution.
Rapid occupancy probability changes together with increasing field of
  view result in more affected voxels, which increases the amount of data that is
transmitted. Voxel occupany probability is unclamped even if
thresholds are exceeded. Similar to the LiDAR reconstruction, the MCG approach has
lower average reconstruction error as compared to the OG approach
(see \cref{sfig:dense_voxel_tof}) as shown in~\cref{tab:reconstruction_accuracy}.

\subsection{Hardware Experiments \label{ssec:hardware_results}}
\subsubsection{Visual-Inertial Navigation and Control\label{ssec:vins}}
State estimates are computed from IMU and downward-facing camera observations via
VINS-Mono~\citep{Qin18_TRO}, a tightly-coupled visual-inertial odometry framework that jointly
optimizes vehicle motion, feature locations, and IMU biases over a sliding
window of monocular images and pre-integrated IMU measurements.
The loop closure functionality of VINS-Mono is disabled to avoid having relocalization-induced
discontinuities in the trajectory estimate, which would have significant implications for
occupancy mapping and is left as future work.

For accurate trajectory tracking, a cascaded Proportional-Derivative (PD)
controller is used with a nonlinear Luenberger observer to compensate for
external acceleration and torque disturbances acting on the
system~\citep{Michael2010}. To improve trajectory tracking, the controller uses
angular feedforward velocity and acceleration terms computed from jerk and snap
references computed from the reference trajectory's $8^{\text{th}}$ order
polynomial (\cref{fig:methodology_overview}).

Additionally, a state machine enables the user to trigger transitions between
the following modes of flight operation: (1) takeoff, (2)
hover, (3) tele-operation, (4) autonomous exploration, and (5) landing. The
results presented in the next section all pertain to the autonomous exploration mode.
In all of the trials included in this paper, the operator only
intervenes to end the trial. For the full duration in all experiments,
the robot operates completely autonomously.


\subsubsection{Implementation Details}
The exploration framework is deployed to the aerial system shown
in~\cref{sfig:rocky0803}, a \SI{2.5}{\kilo\gram} platform equipped with a
forward-facing Intel Realsense D435, downward-facing MV Bluefox2 camera, and
downward and forward facing lights from Cree Xlamp XM-L2 High Power LEDs (Cool
White 6500K). The MV Bluefox2 and D435 cameras operate at \SI{60}{\hertz} and
output images of size $376 \times 240$ and $848 \times 480$, respectively.  The
MV Bluefox2 images are used in state estimation and the D435 depth images are
throttled to \SI{6}{\hertz} for the the mapping system. The D435 camera
estimates depth by stereo matching features in left- and right-infrared camera images
augmented through a dot pattern projected by an IR projector. The laser power
of the IR projector on the D435 is increased from a default value of
\SI{150}{\milli\watt} to \SI{300}{\milli\watt} in order to improve observation
quality in darkness.

The robot is equipped with an Auvidea J120 carrier board with NVIDIA TX2 and
Gigabyte Brix 8550U that communicate over ethernet. The TX2 performs state
estimation and control functions while the Brix performs mapping and planning.
The flight controller used for all experiments is the Pixracer, but the
platform is also equipped with a Betaflight controller as a secondary flight
controller. Switching between the two controllers can be done via a switch on
the RC transmitter. The drone frame is an Armattan Chameleon Ti LR 7'' on which
a Lumenier BLHeli\_32 32bit 50A 4-in-1 electronic speed controller is mounted.
The aerial system is a quadrotor that uses T-Motor F80 Pro 1900KV motors and DAL Cyclone
7056C propellers.

For all hardware experiments, $\| \maxvelocity \| =
\SI[per-mode=symbol]{0.5}{\meter\per\second}$, $\verticalvelocity =
\SI[per-mode=symbol]{0.25}{\meter\per\second}$, and the motion primitives with
duration $2\tau$ in \cref{stab:tof} are disabled.

\subsubsection{Flight Arena Experiments\label{ssec:flight_arena}}
Experiments were conducted in a flight arena with the aerial system.
Each approach (MCG and OG) was flown five times for
\SI{150}{\second}. The results are shown in~\cref{fig:flight_arena}
and a video of one MCG trial with the live map displayed on the base
station may be found at \url{https://youtu.be/egwjv7YwHPE}.

\Cref{sfig:fa_all_entropy} illustrates the map entropy over time
for the 10 trials (5 for each approach)
and~\cref{sfig:fa_mean_entropy} provides the mean and standard
error for each trial.  The communications plots shown
in~\cref{sfig:fa_all_comms} and ~\cref{sfig:fa_mean_comms} illustrate
the theoretical cost to transmit the data (labeled as \emph{Th. OG}
and \emph{Th. MCG}) that was calculated during the simulation trials
(shown in~\cref{sfig:6c,sfig:6f}) as well as the actual transmitted
data (labeled as \emph{Ac. OG} and \emph{Ac. MCG} in dashed lines). The actual transmitted data is calculated as the cumulative sum
of the size of the UDP packets sent over the WiFi router to the base station.
This plot demonstrates that the theoretical estimate closely
matches the actual transmitted amount of data.
No communications dropouts occurred in these trials because the base
station and router were stationary and close to the aerial system's
flight volume.
~\Cref{ssec:hand_carry} analyzes the effect of the robot
moving away from the base station and router to force a communications
dropout.

\Cref{sfig:fa_flying_robot} contains still images of the flight and
\cref{sfig:fa_reconstructed_map} depicts a live map.~\cref{sfig:planning_times}
plots the number of feasible actions available to the robot at a given
time on the left and the time to plan on the right.

\Cref{sfig:octomap_comparison} plots the theoretical data
transferred and compares it to the OctoMap volumetric
map~\cite{hornung2013octomap}. To generate this plot, a
\SI{0.2}{\meter} leaf size OctoMap that matches the OG approach's voxel size
is created. For each scan added to the map, the set of voxels whose
occupancy values (note: not state) have changed are used to generate an
OctoMap submap (i.e., a new OctoMap that contains the change set).
Submaps for this analysis
are created by storing the incremental change set as an OctoMap to
enable the exact map to be reconstructed on the receiving computer.
The full probabilistic model is saved to disk because occupancy
probabilities must be preserved to enable calculation of the
mutual information~\cite{Zhang2019} for information-theoretic exploration.
The serialized stream does not contain any 3D coordinates.
Instead, the spatial relationships between the nodes are stored in the encoding.
Eight bits per node are used to specify whether a child node exists and
an additional floating point number stores the occupancy value for that node.
Due to this design there is some overhead to encode the multi-resolution
nature of the data structure.
The results demonstrate that the MCG approach also outperforms the OctoMap
approach in terms of communication efficiency.

\Cref{tab:fa_planning} provides statistics for the planning and
mapping components for the MCG trials. The planning times and statistics about the
planning actions are provided. Out of all the trials, the planning subsystem
triggers the stopping action only once (in Trial 0). Timing results for the generated
occupied and free GMMs as well as the time to reconstruct occupancy
are provided.

\Cref{tab:fa_mem_compute} provides mean memory and compute statistics
for the autonomy subsystem, which consists of mapping and planning. It
also provides statistics for the Realsense, Bluefox, and PX4
subsystems to quantify the memory and CPU utilization to stream images
and IMU measurements as well as transmitting cascaded commands to the
flight controller.
Statistics are also provided for communication and state estimation.

\subsubsection{Communication Dropout Experiment\label{ssec:hand_carry}}
A communication dropout was triggered by performing an experiment
  where the aerial system is carried away from the WiFi router until
  it is out of range.~\Cref{sfig:ghc_data_sent,sfig:ghc_data_rec}
depict the data sent from the aerial system and received by the
base station over time (note: the clocks on the aerial system and base
station are not synchronized). A view of the operating environment and map
is shown in~\cref{sfig:ghc_resulting_map}. While data is dropped
(i.e., the packets are lost due to the UDP communications protocol)
at around \SI{35}{\second}, the communications are re-established at around \SI{55}{\second} when the
robot reapproaches the base station.
A video at \url{https://youtu.be/UVn2BbMQRJg} illustrates the experiment.


\begin{figure*}
  \subfloat[\label{sfig:rapps_hardware_entropy}]{%
    \ifthenelse{\equal{\arxivmode}{true}}%
    {\includegraphics[]{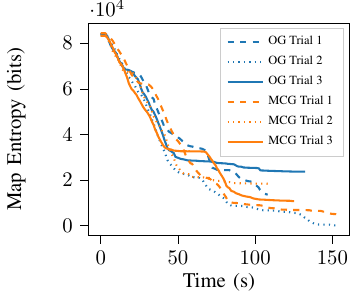}}%
    {\input{./rapps_cave/figures/individual_entropy_trials.tex}}%
  }%
  \subfloat[\label{sfig:rapps_hardware_comms}]{%
    \ifthenelse{\equal{\arxivmode}{true}}%
    {\includegraphics[]{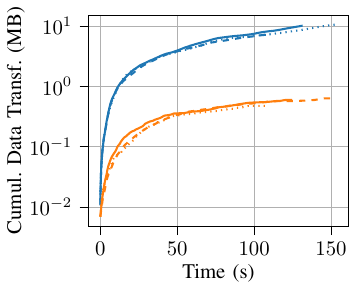}}%
    {\input{./rapps_cave/figures/individual_actual_comms_trials.tex}}%
  }%
  \subfloat[\label{sfig:rapps_cave1}Image courtesy of C. Bassett]{%
    \ifthenelse{\equal{\arxivmode}{true}}%
    {\includegraphics[width=0.3\linewidth,trim=13 0 0 0,clip]{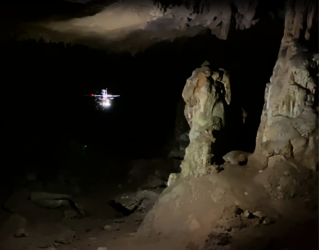}}%
    {\includegraphics[width=0.3\linewidth]{./rapps_cave/images/rapps8.eps}}%
  }\\
  \subfloat[\label{sfig:rapps_cave2}]{%
    \ifthenelse{\equal{\arxivmode}{true}}%
    {\includegraphics[width=\linewidth,trim=0 0 0 0,clip]{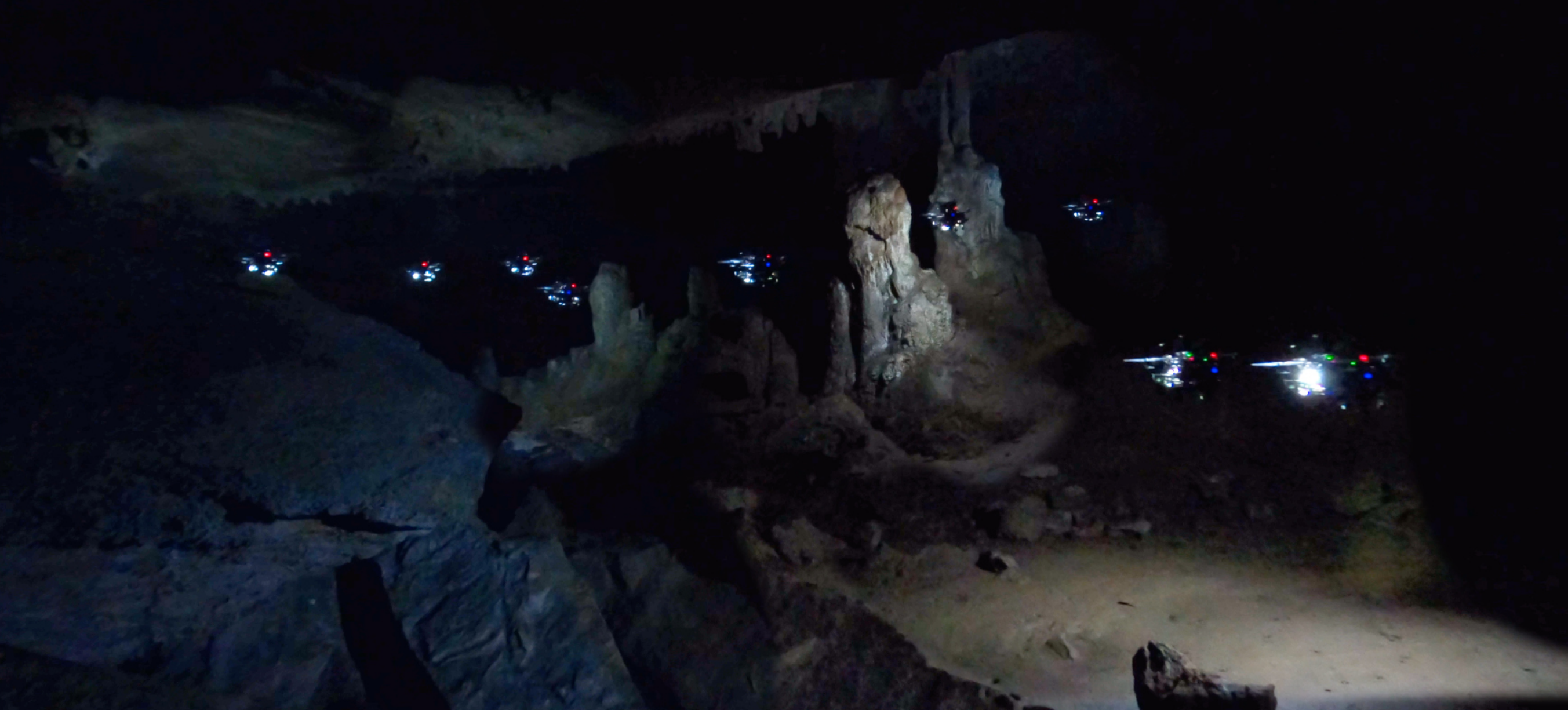}}%
    {\includegraphics[width=\linewidth,trim=0 0 0 0,clip]{./rapps_cave/images/rapps12.eps}}%
  }
\caption{\label{fig:rapps_cave_tests}Overview of the results
from experiments in a cave in West
Virginia.~\protect\subref{sfig:rapps_hardware_entropy} The map entropy over time for three trials of the MCG and OG approaches.
~\protect\subref{sfig:rapps_hardware_comms} The data transferred between a robot and base station for each trial. The communication reported is actual transmitted data over UDP to a base station. Note that while
the exploration performance is similar for both approaches, the data transferred for the MCG approach is substantially less.
~\protect\subref{sfig:rapps_cave1} A still image of the robot flying near a formation with terminated growth.
~\protect\subref{sfig:rapps_cave2} A composite image of one exploration trial composed of still images.}
\vspace{-0.5cm}
\end{figure*}


\subsubsection{Laurel Caverns\label{ssec:laurel_caverns}}
The approach is tested in total darkness at Laurel
Caverns\footnote{http://laurelcaverns.com/}, a commercially operated
cave system in Southwestern Pennsylvania consisting of over four miles
of passages\footnote{The authors acknowledge that caves are fragile
  environments formed over the course of tens of thousands to millions
  of years. Laurel Caverns was chosen as a test site because it has
  relatively few speleothems\footnotemark due to its sandstone overburden and the
  high silica content of the Loyalhanna limestone~\cite{patrick2004pennsylvania}.
The authors worked with cave management to select a test site that
contained low speleothem growth to minimize risk of damage to the
cave. Cave management monitored all flights. No flights were
executed near delicate formations.}\footnotetext{Speleothems
are mineral formations found in limestone caves (e.g., stalagmites,
stalagtites, and flowstone) that are composed of calcium carbonate,
precipated from groundwater that has percolated through adjacent
carbonate host rock~\cite{speleothems}.}\footnote{Bat populations in
the northeastern U.S. have been decimated with the onset of White-nose
Syndrome in the winter of 2007-2008~\cite{frick2010emerging}. Great
care was taken not to disturb bats with the aerial systems during the
hibernating season.}.

\Cref{sfig:rocky0803_explores} illustrates a composite image from
several still images of the robot exploring the Laurel Caverns Dining
Room. Two experiments were conducted, one for each of the MCG and OG
approaches for a $\SI{95}{\second}$ duration. The map entropy
reduction over time is shown in \cref{sfig:laurel_caverns_entropy} and
is similar for both approaches, while the cumulative data transferred
(\cref{sfig:laurel_caverns_comms}) to represent the maps is
more than an order of magnitude lower for the MCG approach as
compared to the OG approach. Note, however, that the communication
reported for this experiment represents the theoretical, or estimated,
communications needed to transmit the data. The data was not transmitted to a
base station. The data transfer rate in~\cref{sfig:laurel_caverns_rate}
is calculated using Euler differentiation but note that the accuracy is
affected by the limited number of samples. During hardware trials, a bounding
box was used to constrain the exploration volume.
To put the localization accuracy into perspective, the drift
in position is about $\SI{0.53}{\meter}$ during a $\SI{50.9}{\meter}$ cave flight
and the rotation drift is about $\SI{0.32}{\radian}$ over $\SI{33.5}{\radian}$
which is about a 1\% drift in both translation and rotation.
Position drift may be approximated as the difference between the initial
and final position estimates because the robot takes off and lands at the
same location.


\subsubsection{West Virginia Cave\label{ssec:rapps_cave}}
The approach was also tested in total darkness in a cave in West
Virginia\footnote{The region of the cave where flight experiments were
conducted contained speleothems that have ceased growing.  Speleothems
growth may terminate due to geologic, hydrologic, chemical, or
climatic factors that cause water percolation to cease at a particular
drip site~\cite{speleothems}. The authors worked with cave management
to select a test site that had neither actively growing speleothems or
bats.}.~\Cref{sfig:rapps_hardware_entropy} illustrates the map entropy
reduction over time with a maximum duration of \SI{150}{\second}.
The MCG and OG approaches perform similarly in these trials.
The actual data transferred between the robot and base station
is shown in~\cref{sfig:rapps_hardware_comms}.~\Cref{sfig:rapps_cave1} is a still image of the robot approaching
a formation and~\cref{sfig:rapps_cave2} is a composite image from
several still images of the robot exploring the cave.
A video of one exploration run may be found at the following link: \url{https://youtu.be/H8MdtJ5VhyU}.
In these experiments a bounding box was used to constrain the exploration volume.
The drift in position is about \SI{1.28}{\meter}
during the \SI{82.6}{\meter} flight and rotation drift is about
\SI{0.55}{\radian} over \SI{41.8}{\radian} which is about a 1.5\%
drift in position and 1.3\% drift in rotation.



\section{Discussion and Limitations\label{sec:discussion}}
A limitation of the current work is that while the simulation
  assumes perfect state estimates, the VINS state estimator used onboard
  the aerial robot drifts over time. While it is not possible
  to entirely eliminate noise, some ways to mitigate the error
  are by (1) incorporating the depth observations from the forward-facing
  realsense camera into the VINS estimates and (2) incorporating
  loop closures.

Another limitation of the work is multiple representations are
required to enable mapping, planning and state estimation.  The map
representation leverages GMMs, planning utilizes occupancy grid maps,
and state estimation uses a sparse-point set.

Finally, there are many parameters that must be tuned in order
for the system to function well. A potential source of future
work is to automate this process.


\section{Conclusions and Future Work\label{sec:conclusion}}
The results presented in this paper comprise the beginning of a
promising line of research for autonomous cave surveying and mapping
by aerial systems. A high-fidelity model amenable to
transmission across low-bandwidth communications channels is achieved
by leveraging GMMs to compactly represent the environment.
The method is demonstrated with \SI{360}{\degree} and limited field of view
sensors utilizing a planning framework that is amenable to both sensor models.
Several avenues of future work remain.
Cave maps are typically annotated with important terrain
features such as stalactites, stalagmites, and breakdown
(see~\cref{fig:sf_map} for example), but this
work does not consider the problem of terrain feature classification
and encoding. Additionally, waterproof, rugged, and easy-to-use 2D maps
are critical for cave
rescuers or explorers to avoid getting lost in caves.
Methods to project the 3D map information from the robot
to 2D are needed to fill this gap in the state of the art.
Finally, the deployment of
multiple robots to increase the speed of exploration is of interest
for large passages or maze caves. Introducing re-localization strategies to curb drift over long
duration flights and yield more consistent maps would also
be beneficial for multi-robot operations. Beyond cave
applications, this work has relevance for search
and rescue, planetary exploration, and tactical operations where
humans and robots must share information in real-time.


\section{Acknowledgments}
The authors thank C. Bassett for faciliating experiments at
the West Virginia cave. The authors also thank H. Brooks and
R. Maurer for facilitating experiments at Laurel Caverns and thank
D. Cale for granting permission to test at Laurel Caverns. The
authors also thank D. Melko for support and guidance regarding
test sites, lending equipment and teaching the authors about
caving. The authors thank B. Ashbrook for his insights and
information regarding the cave on the Barbara Schomer Cave Preserve in
Clarion County, PA. The authors thank H. Wodzenski and J. Jahn
for providing images used in this work. Finally, the authors thank
X. Yang for fruitful discussions about motion primitives-based
planning and A. Dhawale, A. Desai, E. Cappo, T. Lee,
M. Collins, and M. Corah for feedback on this manuscript.

\balance
\bibliographystyle{IEEEtranN}
{
  \footnotesize
  \bibliography{refs}
}

\begin{IEEEbiography}[{\includegraphics[width=1in,height=1.25in,clip,keepaspectratio]{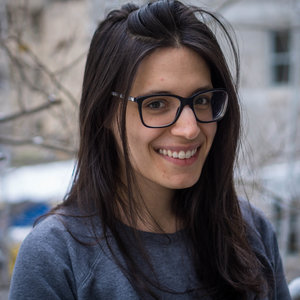}}]{Wennie Tabib}
  received the B.S. degree in computer science in 2012, the
M.S. degree in robotics in 2014, and the Ph.D. degree in computer
science in 2019 from Carnegie Mellon University, Pittsburgh, PA,
USA.

She is currently a Systems Scientist with the Robotics Institute at Carnegie
Mellon University. She researches perception, planning, and learning
algorithms to enable safe autonomy in significantly three-dimensional,
complex environments. Her current research develops methods to enable
aerial systems to explore subterranean environments. Wennie is also a
member of the National Speleological Society (NSS 69985), Mid-Atlantic
Karst Conservancy, Pittsburgh Grotto and Loyalhanna Grotto.
\end{IEEEbiography}
\begin{IEEEbiography}[{\includegraphics[width=1in,height=1.25in,clip,keepaspectratio]{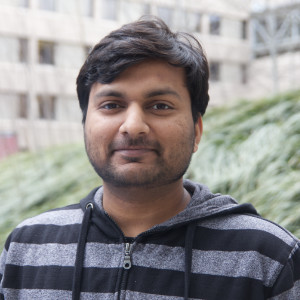}}]{Kshitij Goel}
  received the B.Tech. degree in aerospace engineering in 2017 from the Indian
  Institute of Technology (IIT) Kharagpur, Kharagpur, WB, India.

  He is currently a Ph.D. student in Robotics at Carnegie Mellon University, researching fast motion planning for
multirotors operating in unknown environments. His current work focuses on
robustly deploying teams of multirotors to rapidly explore challenging real world scenarios in real time.
\end{IEEEbiography}
\begin{IEEEbiography}[{\includegraphics[width=1in,height=1.25in,clip,keepaspectratio]{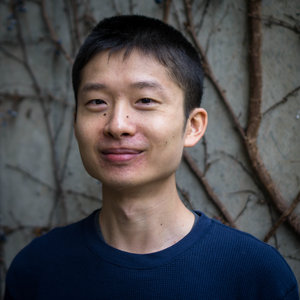}}]{John Yao}
  received the B.A.Sc. degree in aerospace engineering from the
University of Toronto, Toronto, Canada, in 2013 and the M.S. degree in robotics
from Carnegie Mellon University (CMU), Pittsburgh, PA, USA in 2016.

John is a Ph.D. Candidate in the Robotics Institute at CMU.  His
research interests include visual-inertial state estimation and
resource-constrained sensor fusion for autonomous robots.
\end{IEEEbiography}
\begin{IEEEbiography}[{\includegraphics[width=1in,height=1.25in,clip,keepaspectratio]{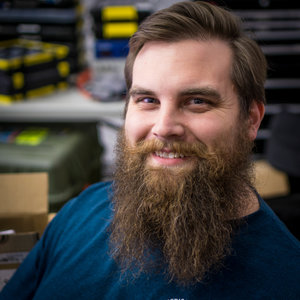}}]{Curtis Boirum}
  received the B.S. degree in physical science from Eureka College,
Eureka, IL, USA, in 2008.  He received the B.S. and M.S. degrees in
mechanical engineering from Bradley Unversity, Peoria, IL, USA, in
2009 and 2011, respectively.  He received the M.S. degree in robotics
from Carnegie Mellon University, Pittsburgh, PA, USA, in 2015.

  Curtis is a systems engineer for the Resilient Intelligent Systems
Lab who designs, builds, and operates drones and ground robots ranging
in size from 100g to 7kg.
\end{IEEEbiography}
\begin{IEEEbiography}[{\includegraphics[width=1in,height=1.25in,trim=50 50 50 50,clip,keepaspectratio]{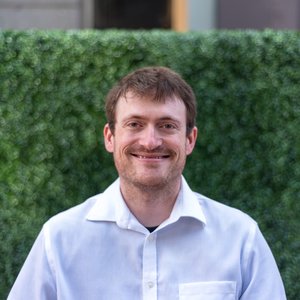}}]{Nathan Michael}
  received the Ph.D. degree in mechanical engineering from the
University of Pennsylvania, Philadelphia, PA, USA, in 2008.

  He is an Associate Research Professor in the Robotics Institute of
Carnegie Mellon University; Director of the Resilient Intelligent
Systems Lab; author of over 160 publications on control, perception,
and cognition for resilient intelligent single and multi-robot
systems; nominee or recipient of nine best paper awards; recipient of
the Popular Mechanics Breakthrough Award and Robotics Society of Japan
Best Paper Award (of 2014); PI of past and ongoing research programs
supported by ARL, AFRL, DARPA, DOE, DTRA, NASA, NSF, ONR, and
industry; and Chief Technical Officer of Shield AI. Nathan develops
resilient intelligent autonomous systems capable of individual and
collective intelligence through introspection, adaptation, and
evolvement in challenging domains.
\end{IEEEbiography}


\end{document}